\documentclass[conference]{IEEEtran}
\IEEEoverridecommandlockouts
\usepackage{cite}
\usepackage{amsmath,amssymb,amsfonts}
\usepackage{algorithmic}
\usepackage[vlined,ruled,noend,linesnumbered]{algorithm2e}
\usepackage{graphicx}
\usepackage{subfigure}
\usepackage{textcomp}
\usepackage{xcolor}
\usepackage{hyperref}
\hypersetup{colorlinks=true,linkcolor=blue,citecolor=blue,urlcolor=blue}
\usepackage{tikz}

\def\BibTeX{{\rm B\kern-.05em{\sc i\kern-.025em b}\kern-.08em
    T\kern-.1667em\lower.7ex\hbox{E}\kern-.125emX}}
\begin{document}

\title{Evolving Benchmark Functions to Compare Evolutionary Algorithms via Genetic Programming}


\author{\IEEEauthorblockN{Yifan He}
\IEEEauthorblockA{
\textit{Zhejiang University of Finance \& Economics}\\
Hangzhou, China \\
yifanhe@zufe.edu.cn}
\and
\IEEEauthorblockN{Claus Aranha}
\IEEEauthorblockA{
\textit{University of Tsukuba}\\
Tsukuba, Japan \\
caranha@cs.tsukuba.ac.jp}
}

\maketitle

\begin{abstract}

In this study, we use Genetic Programming (GP) to compose new optimization benchmark functions. Optimization benchmarks have the important role of showing the differences between evolutionary algorithms, making it possible for further analysis and comparisons.  We show that the benchmarks generated by GP are able to differentiate algorithms better than human-made benchmark functions. The fitness measure of the GP is the Wasserstein distance of the solutions found by a pair of optimizers. Additionally, we use MAP-Elites to both enhance the search power of the GP and also illustrate how the difference between optimizers changes by various landscape features. Our approach provides a novel way to automate the design of benchmark functions and to compare evolutionary algorithms.
\end{abstract}


\newcommand\copyrighttext{%
  \footnotesize\centering\color{blue} \copyright 2024 IEEE. Personal use of this material is permitted. Permission from IEEE must be obtained for all other uses, in any current or future media, including reprinting/republishing this material for advertising or promotional purposes, creating new collective works, for resale or redistribution to servers or lists, or reuse of any copyrighted component of this work in other works.}
\newcommand\copyrightnotice{%
\begin{tikzpicture}[remember picture,overlay]
\node[anchor=south,yshift=10pt] at (current page.south) {{\parbox{\dimexpr\textwidth-\fboxsep-\fboxrule\relax}{\copyrighttext}}};
\end{tikzpicture}%
}

\copyrightnotice

\section{Introduction}
\label{sec:introduction}


Optimization benchmarks are mathematical functions that represent optimization problems. They are used to compare different evolutionary algorithms, to analyze their behavior, to fine tune their parameters, and also as starting points to create new algorithms that are effective against specific characteristics of an optimization problem. The automatic design of Evolutionary Algorithms also requires good benchmark functions to compare algorithm instances.

Good benchmark functions represent characteristics of different families of real-world problems, such as modality, ruggedness, etc. Evolutionary algorithms designed on these benchmarks are supposed to work well on the same types of the problems in the real world. Even then, many times it is not possible to predict the characteristics of a specific real world problem.

Good benchmark functions can also be used to separate or differentiate between existing evolutionary algorithms. 
However, to generate such functions is rather challenging. It requires expert level knowledge about the algorithms to be compared, as well as mathematical knowledge to compose optimization functions with specific characteristics. 

\begin{figure}[t]
    \centering
    \includegraphics[scale=.15]{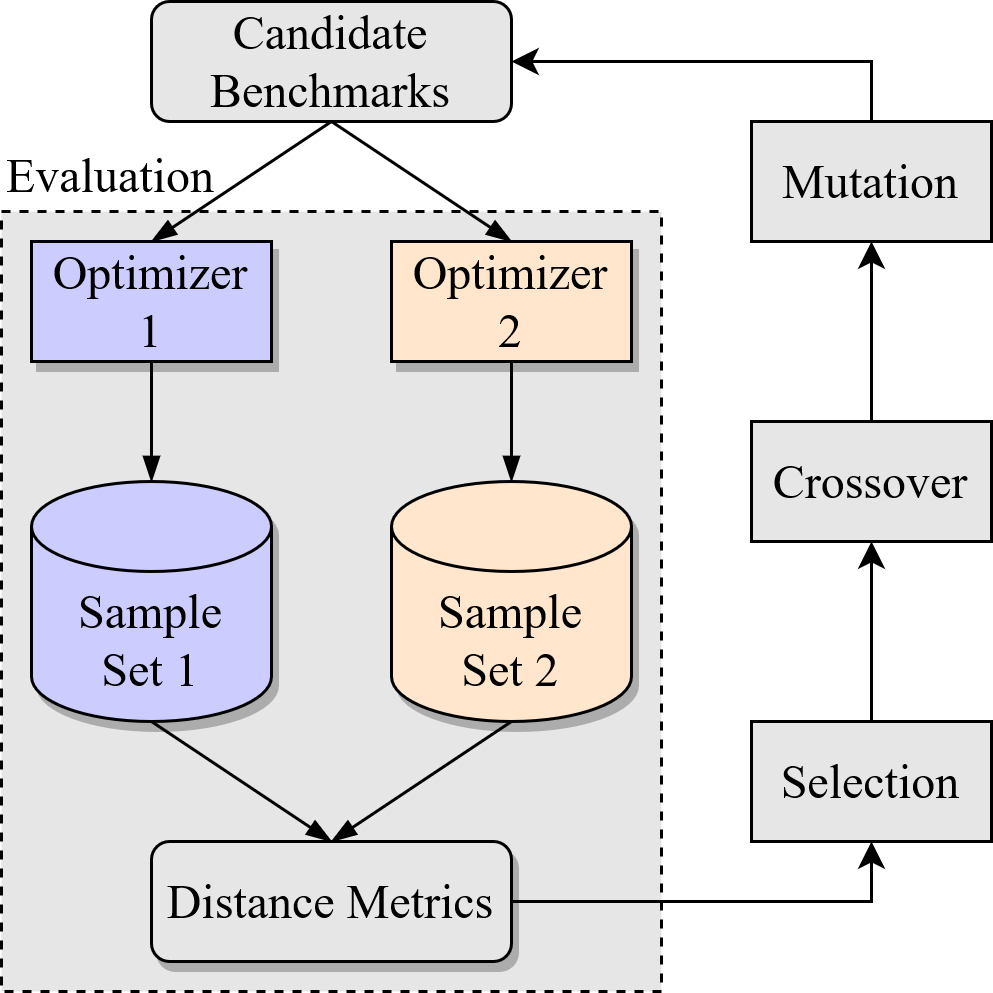}
    \caption{General outline of the proposed GP system}
    \label{fig:general-idea}
\end{figure}

In this paper, we propose using Genetic Programming (GP)~\cite{koza1994genetic} as a solution to this interesting but challenging task. We use GP to create a set of benchmark functions. These functions are selected by their ability to differentiate between two or more algorithms. In this initial research, we focus on the two-algorithm case. Our approach is illustrated in Fig.~\ref{fig:general-idea}. The key component of this approach is the evaluation of the generated functions. We apply two optimizers to find solutions to the candidate function, and measure the distance between the solution sets obtained from each optimizer. In this research, we use multidimensional Wasserstein distance~\cite{kantorovich1960mathematical} to compute the distance between the parameter distribution of the two sample sets.

Moreover, to find a set of benchmark functions with diverse characteristics, we use MAP-Elites~\cite{mouret2015illuminating} in our GP system. MAP-Elites has proven itself as an effective quality-diversity technique in several difficult problems, and it can also be used to illustrate how the two optimizers are different on a set of functions with different characteristics, such as fitness landscape feature~\cite{malan2021survey}. In our GP system, MAP-Elites is used to find a set of varied benchmark functions with different performance characteristics.

We compare the benchmark functions generated by our GP with functions from the CEC2005 benchmark~\cite{suganthan2005problem}. This comparison is made on two pairs of algorithms, 1) two different configurations of Differential Evolution (DE)~\cite{storn1997differential}, as a proof-of-concept and 2) two powerful optimizers, namely SHADE~\cite{tanabe2013success} and CMA-ES~\cite{hansen2001completely}, which have obtained good performance in the CEC benchmark in the past. 
First the GP system generates a set of 2-dimensional benchmark function (train performance), then we extend these functions to 10-dimensions to validate the result (test performance). 

The functions produced by our approach separate both algorithm pairs better than the CEC2005 benchmarks in the decision space, and hold a comparable performance in differentiating algorithms in the objective space. In addition, the heatmaps of the archives retrieved by MAP-Elites indicate that the two DE configurations are more different when the Fitness Distance Correlation (FDC)~\cite{jones1995fitness} and the neutrality~\cite{reidys2001neutrality} of the function are close to 0, while SHADE and CMA-ES are more different when FDC is larger than -0.2.

We believe that our approach could be further developed into a good tool for generating benchmark functions to analyze the difference between sets of evolutionary algorithms.





\section{Preliminaries}
\label{sec:preliminaries}

\subsection{Optimization Benchmarks}
\label{sec:optimization-benchmarks}

For a new evolutionary algorithm, it is important to compare its performance with existing algorithms on a good set of optimization benchmarks. One common practice is to include a large set of test problems, since the difference between two algorithms, i.e., the effectiveness of one algorithm against others, cannot be measured if the set of problems are too specialized and without diverse properties~\cite{jamil2013literature}. Therefore, a variety of benchmark functions are proposed with different characteristics, such as modality, regularity, separability, and dimensionality. Jamil \textit{et al.}~\cite{jamil2013literature} provided a comprehensive review on over a hundred of existing benchmark functions, such as functions from CEC2005 benchmark~\cite{suganthan2005problem}.

In the first chapter of Ackley's dissertation~\cite{ackley1987connectionist}, he showed his steps of making a multimodal maximization function. He started from a simple unimodal function $z=200\exp{(-0.2\sqrt{x^2+y^2})}$. Based on this unimodal function, he came up with a multimodal function by adding a trigonometric term $5\exp{(\cos{3x}+\sin{3y})}$. Most of the existing benchmarks are created following the similar approach, i.e., starting with simple functions, adding or multiplying some terms, performing transformations, and making composite functions. However, this approach requires mathematical knowledge to create functions. Moreover, many times the composed functions fail to show the difference between powerful algorithms. A review paper~\cite{bartz2020benchmarking} summarized several issues in the current benchmark problems, such as 1) not close to real-world cases and 2) the overfitted algorithm design to the existing benchmarks. These limitations motivate the idea of automatic benchmark function composition.


The idea of evolving problem instances has been explored a few times before. Mu{\~n}oz~\textit{et al.}~\cite{munoz2020generating} applied instance space analysis to an existing benchmark suite and used GP to generate problems with different features from the existing ones. In the study by Long~\textit{et al.}~\cite{long2023challenges}, GP is used to create functions with similar landscape features to a set of existing problems to build surrogate functions. They proposed to minimize the Wasserstein distance between landscape feature vectors of generated functions and target problems using a classical tree-based GP. Lechien~\textit{et al.}~\cite{lechien2023evolving} evolved graph instances to compare traveling salesman problem optimizers.

Unlike the prior studies~\cite{munoz2020generating,long2023challenges}, we focus on functions for comparing and analyzing a specific pair/set of optimizers. Such functions could be used, for instance, to tell when a new algorithmic component is efficient/inefficient. In this way, these functions are important for the development of new algorithms. Therefore, instead of generating diverse functions or surrogate functions, we propose to compose functions that can show difference between similar or powerful evolutionary optimizers. For this purpose, we design a different objective function of GP to maximize the difference between solution sets of two optimizers. We additionally introduce MAP-Elites selection~\cite{mouret2015illuminating} to illustrate how the difference between optimizers changes by various landscape features.




\subsection{Techniques Used}
\label{sec:techniques-used}

In this study, we generate mathematical benchmark functions via GP~\cite{koza1994genetic}. Additionally, we use MAP-Elites~\cite{mouret2015illuminating} with fitness landscape metrics~\cite{malan2021survey} to 
select multiple functions with diverse characteristics out of the population. We introduce the three important techniques we used in the following paragraphs.

\subsubsection{Koza's Genetic Programming}
\label{sec:kozas-genetic-programming}

GP~\cite{koza1994genetic} is an evolutionary algorithm that evolves computer programs based on users' specifications. The initial study of GP by Koza~\cite{koza1994genetic} uses a tree representation of programs where every node takes its children as arguments and the root node returns the final output of the program/tree. These nodes are selected from a given set of functions and terminals. Koza's GP uses tournament selection, subtree crossover, and subtree mutation. A bloat control technique that ignores the offspring with a large tree size is usually applied to prevent the infinite increase of the search space. GP is a common method in generating mathematical functions to fit given datasets, i.e., symbolic regression problems~\cite{mcdermott2012genetic}.

\subsubsection{MAP-Elites}
\label{sec:map-elites}

\begin{algorithm}[t]
\caption{Pseudo-code of MAP-Elites algorithm}
\label{alg:map-elites}
\KwIn{maximum number of evaluation $\mathrm{nfe}$ and empty archive $\mathcal{A}$}
\KwOut{filled archive $\mathcal{A}$}
\While{$t < \mathrm{nfe}$}
{
    \eIf{$|\mathcal{A}|< s$}
    {
        $x \gets \mathrm{random\_initialize}()$\;
    }{
        $p_1, \dots, p_n \gets \mathrm{random\_select}(\mathcal{A})$\;
        $x \gets \mathrm{reproduce}(p_1, \dots, p_n)$\;
    }
    $[v_1,\dots,v_m] \gets \mathrm{phenotype}(x)$\;
    $e \gets \mathcal{A}[v_1,\dots,v_m]$\;
    \If{$e$ is None or $f(x) < f(e)$}
    {
        $\mathcal{A}[v_1,\dots,v_m] \gets x$\;
    }
    $t \gets t+1$\;
}
\Return $\mathcal{A}$\;
\end{algorithm}

The Multi-dimensional Archive of Phenotypic Elites (MAP-Elites)~\cite{mouret2015illuminating} is a quality-diversity algorithm that maintains an archive of elite solutions with diverse phenotype. For this purpose, several phenotypic descriptors are defined at the initialization step. The archive is then divided into partitions based on intervals on phenotypic values. MAP-Elites selects random parents from the archive to generate child individuals. These child individuals are then compared to elites with the same phenotype in the archive. This localized selection ensures the diversity in the pre-defined space by the given phenotypic descriptors. Moreover, the algorithm generates random individuals until a certain percentage of partitions of the archive is filled so that the random parent selection can provide enough diversity. Algorithm~\ref{alg:map-elites} provides the pseudocode of MAP-Elites. MAP-Elites has been applied in many fields, such as machine learning~\cite{ecoffet2021first} and program synthesis~\cite{dolson2019exploring}.

\subsubsection{Fitness Landscape Analysis}
\label{sec:fitness-landscape-analysis}

Fitness Landscape Analysis (FLA)~\cite{malan2021survey} is a group of techniques that quantify the characteristics of optimization problems. These characteristics can influence the algorithmic behavior and eventually contribute to the performance of evolutionary solvers. Fitness Distance Correlation (FDC)~\cite{jones1995fitness} characterizes the deception of the landscape with respect to local search. As in (\ref{eq:fitness-distance-correlation}), FDC takes random samples from the search space and computes the distances $d_i$ between these samples to the optimal solution, as well as their fitness values $f_i$. After that, the correlation $r$ between these distances and fitness values are computed to quantify the deception of the landscape. A small FDC indicates a high deception in the landscape.
\begin{gather}
\label{eq:fitness-distance-correlation}
    r = \frac{C_{\mathrm{FD}}}{{S_F}{S_D}}\\
    C_{\mathrm{FD}} = \frac{1}{n} \Sigma_{i=1}^n (f_i-\bar{f})(d_i-\bar{d})
\end{gather}

Neutrality~\cite{reidys2001neutrality} is another important FLA metric that describes the area with no (or little) change in fitness. The individuals in highly neutral areas are hard to evolve. The neutrality is computed as in (\ref{eq:neutrality}). A random walk of $T$ steps is performed and the rate of neighbor solutions that share the same fitness is calculated. $\mathcal{I}_\varepsilon(\cdot)$ is a function that returns 1 when the difference between two arguments is less than a small positive value $\varepsilon$ and 0 otherwise.
\begin{equation}
\label{eq:neutrality}
    p = \frac{\Sigma_{i=1}^{T-1} \mathcal{I}_{\varepsilon}(f_i, f_{i+1})}{T-1}
\end{equation}

\section{Proposed Method}
\label{sec:proposed-method}
We propose a GP method to generate benchmark functions. The GP evolves functions that highlight differences in performances between a pre-selected pair of algorithms.

\subsection{Function Set and Terminals}
\label{sec:function-set-and-terminals}

\begin{table}[]
\caption{Function set and terminals used in our GP}
\label{tab:function-set-and-terminals}
\centering
\begin{tabular}{|c|c|}
    \hline
    \textbf{Category} & \textbf{Elements} \\
    \hline
    Operators & add, sub, mul, neg, sqrt, sin, cos \\
    Terminals & variables ($x_0, \dots, x_D$) \\
    Constants & integers in $[-10,10]$ \\
    \hline
\end{tabular}
\end{table}

The function set and terminals used 
are listed in Table~\ref{tab:function-set-and-terminals}. We do not use operators such as ``protected division'' that return non-continuous outputs.

\subsection{Evaluation Metrics}
\label{sec:evaluation-metrics}


Our proposed method aims to find one or more benchmark functions that maximize the difference of the behavior between two optimizers. We define the behavior of an optimizer as the parameter distribution of all solutions it sampled during $n$ repetitions of the evolution process. We use Wasserstein distance~\cite{kantorovich1960mathematical} to measure the difference of the parameter distributions between two solution sets. The Wasserstein distance can be seen as the minimum amount of work to transform one distribution into another. We use the default implementation in the SciPy library. For the multidimensional distributions, we compute the Wasserstein distance in every dimension and take the average as shown in (\ref{eq:wasserstein-distance}). $\mathcal{U}_i$ and $\mathcal{V}_i$ are two sets of parameters in the $i$-th dimension. A larger difference returns a bigger value of $d$. The pseudocode of the distance calculation is provided in Algorithm~\ref{alg:distance-measure}.
\begin{equation}
\label{eq:wasserstein-distance}
    d = \frac{1}{D} \Sigma_{i=1}^D d_{w}(\mathcal{U}_i, \mathcal{V}_i)
\end{equation}

\subsection{MAP-Elites using Landscape Metrics}
\label{sec:map-elites-using-landscape-metrics}

\begin{figure}
    \centering
    \includegraphics[scale=.15]{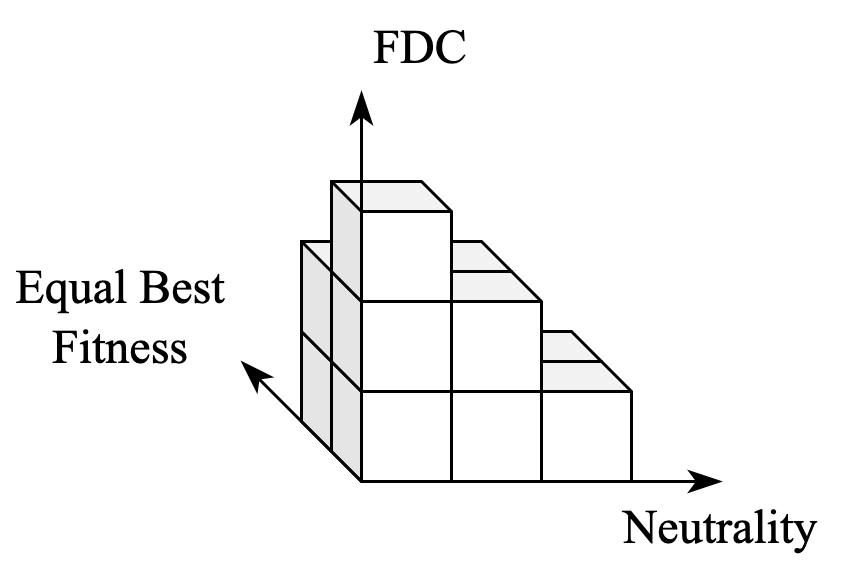}
    \caption{Phenotypic descriptor of the MAP-Elites in our study}
    \label{fig:3-dimensional-archive}
\end{figure}

When generating multiple benchmark functions, we want them to be different enough that they examine different characteristics of the algorithms under studies. To achieve this, we add MAP-Elites to the proposed GP, so that the selection step in the GP can maintain candidate functions with different characteristics, represented as the different phenotypic descriptions (metrics) of MAP Elites.

In this study, we use FDC and neutrality as phenotypic descriptors. The calculation of these two metrics has been introduced in Section~\ref{sec:fitness-landscape-analysis}. We further introduce another phenotypic descriptor that indicates whether the optimization result, i.e., the best fitness value, of the two optimizers on the function is the same or not. Therefore, in our proposed method, the MAP-Elites maintains a three-dimensional archive (shown in Fig.~\ref{fig:3-dimensional-archive}).

Unlike the original MAP-Elites algorithm~\cite{mouret2015illuminating}, our implementation generates a small population of child individuals at each step of the loop to exploit the power of parallel computing. We provide pseudocode of our method in Algorithm~\ref{alg:proposed-method} and Python implementation in an online repository at \url{https://github.com/Y1fanHE/cec2024}.

\begin{algorithm}[t]
\caption{Pseudocode to of distance evaluation}
\label{alg:distance-measure}
\KwIn{mathematical function $h$, two evolutionary optimizers $\mathrm{opt}_1$ and $\mathrm{opt}_2$, and repetition $n$}
\KwOut{distance metric value $d$}
$\mathcal{U} \gets \emptyset, \mathcal{V} \gets \emptyset$\;
\For{$k \in \{1,\dots,n\}$}
{
    $X \gets \mathrm{random\_initialize}()$\;
    $\mathcal{U}_k \gets \mathrm{opt}_1(h, X), \mathcal{V}_k \gets \mathrm{opt}_2(h, X)$\;
    $\mathcal{U} \gets \mathcal{U} \cup \mathcal{U}_k, \mathcal{V} \gets \mathcal{V} \cup \mathcal{V}_k$\;
}
$d \gets \mathrm{wasserstein\_distance}(\mathcal{U},\mathcal{V})$\;
\Return $d$\;
\end{algorithm}

\begin{algorithm}[t]
\caption{Pseudocode of the proposed method}
\label{alg:proposed-method}
\KwIn{maximum number of evaluation $\mathrm{nfe}$, population size $N$, empty archive $\mathcal{A}$, and optimizers to differentiate $\mathrm{opt}_1$ and $\mathrm{opt}_2$} 
\KwOut{filled archive $\mathcal{A}$}
\While{$t < \mathrm{nfe}$}
{
    \eIf{$|\mathcal{A}| < s$}
    {
        \For{$i \in \{ 1,\dots,N \}$}
        {
            $h_i \gets \mathrm{random\_initialize}()$\;
        }
    }{
        $p_1, \dots, p_N \gets \mathrm{random\_select}(\mathcal{A}, N)$\;
        \For{$i \in \{ 1,\dots,\frac{N}{2} \}$}
        {
            $h_{2i}, h_{2i-1} \gets \mathrm{crossover}(p_{2i-1}, p_{2i})$\;
            $h_{2i} \gets \mathrm{mutate}(h_{2i})$\;
            $h_{2i-1} \gets \mathrm{mutate}(h_{2i-1})$\;
        }
    }
    \For{$i \in \{ 1,\dots,N \}$}
    {
        $d(h_i) \gets \mathrm{wasserstein\_distance}(h_i, \mathrm{opt}_1, \mathrm{opt}_2)$\;
        $[v_1,v_2,v_3] \gets \mathrm{phenotype}(\mathbf{x}_i)$\;
        $e_i \gets \mathcal{A}[v_1,v_2,v_3]$\;
        \If{$e_i$ is None or $d(h_i) < d({e}_i)$}
        {
            $\mathcal{A}[v_1,v_2,v_3] \gets h_i$\;
        }
    }
    $t \gets t+N$\;
}
\Return $\mathcal{A}$\;
\end{algorithm}

\section{Experiments}
\label{sec:experiments}

Initially, we focus on generating two-dimensional functions with boundary of $[-5,5]$. To evaluate a function, we use two algorithms (described in the case studies) to find solutions for the candidate function, with three repetitions. The average of the Wasserstein distance between the algorithms' solution archives in the three runs is used as the quality measure.

To compute the FDC of the candidate function, we take 5000 samples randomly. To compute the neutrality, we perform 5000 steps of a random walk. $\varepsilon$ is set to 0.005, that is, a pair of neighbor solution is neutral if the difference between their fitness is less than 0.005. We divide FDC and neutrality into 20 partitions, respectively. As a result, the MAP-Elites archive has 20$\times$20$\times$2=800 partitions (the last ``2'' being for the third phenotypic descriptor mentioned in Section~\ref{sec:map-elites-using-landscape-metrics}).

In the GP, we use a population size of 50 and maximum generation of 1000. We tune the important parameters as follows. We first tune the combination of crossover rate and mutation rate from $\{0.6,0.7,0.8,0.9,1.0\}\times\{0.0,0.1,0.2,0.3,0.4\}$, with a fixed value of MAP-Elites parameter $s=200$. We then tune MAP-Elites parameter $s$ from $\{100,200,300,400\}$ with the tuned values of crossover rate and mutation rate. We select the parameters based on how they differentiate between two algorithm configurations in the first case study (see Section~\ref{sec:case-study-on-parameterization-of-differential-evolution}). Consequently, the crossover rate and the mutation rate are set to 0.9 and 0.3, respectively. The MAP-Elites parameter $s$ is set to 200, i.e., the algorithm generates individuals by random initialization rather than reproduction steps until 200 partitions in the archive are filled. The tree height in the initialization step is in $[3,6]$ and the maximum tree height during the reproduction is set to 10. 

\subsection{Case Study on Parametrization of Differential Evolution}
\label{sec:case-study-on-parameterization-of-differential-evolution}

\begin{figure*}[]
    \centering
    \includegraphics[width=.85\columnwidth]{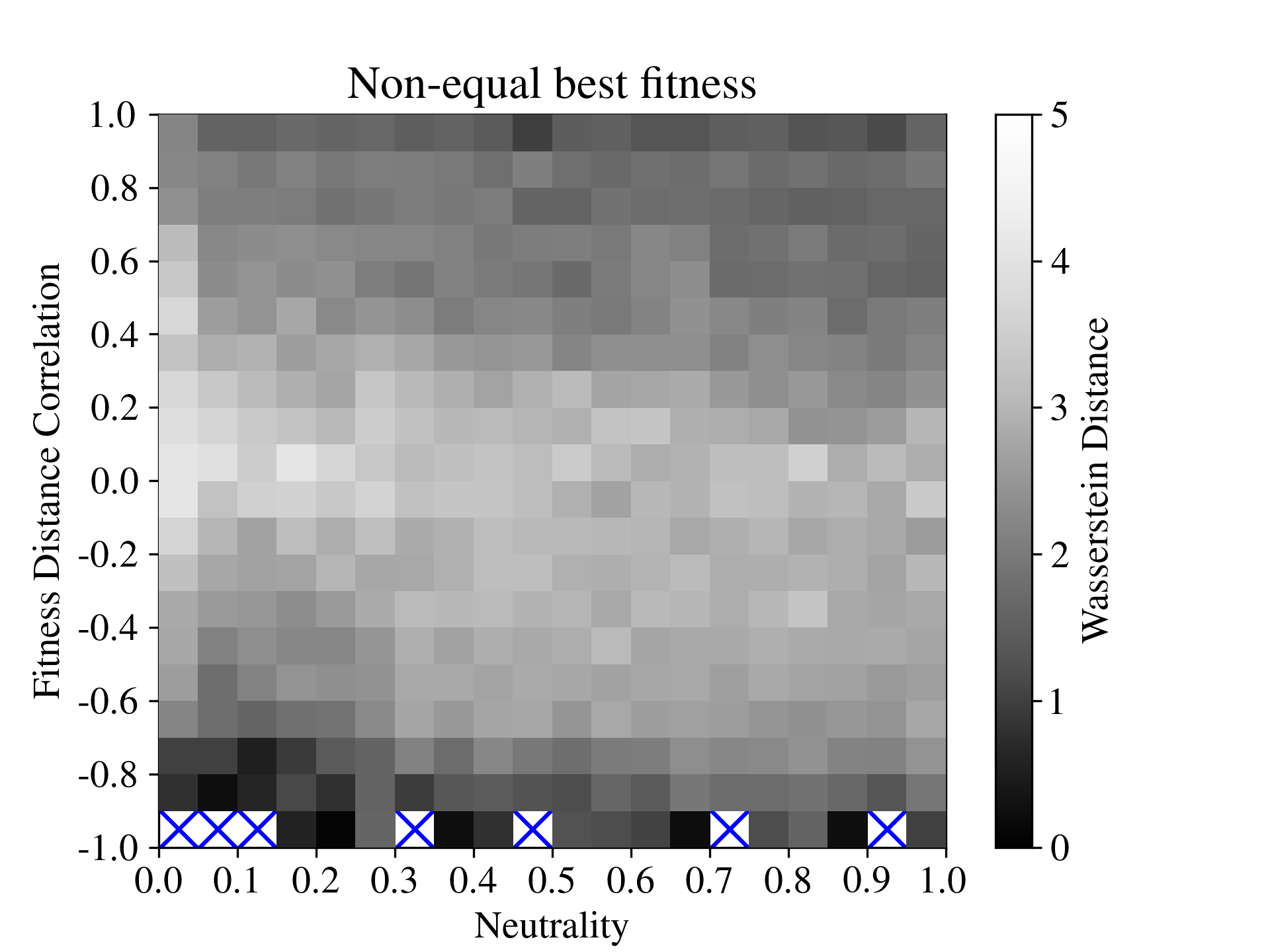}
    \includegraphics[width=.85\columnwidth]{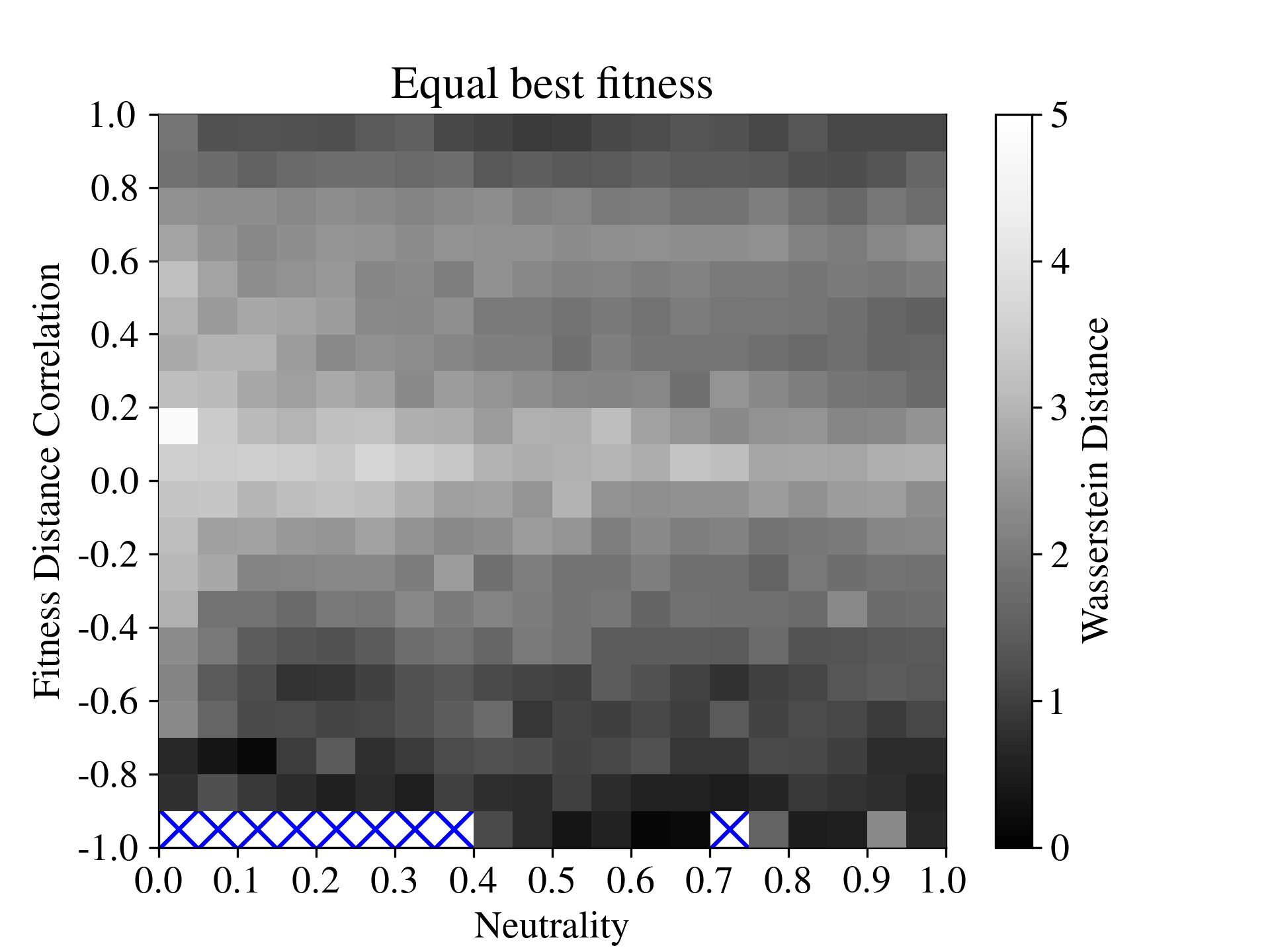}
    \caption{Average Wasserstein distance of the functions in the archive in case study 1}
    \label{fig:archive-de-parameterization}
\end{figure*}

\begin{figure*}[]
    \centering
    \includegraphics[width=.85\columnwidth]{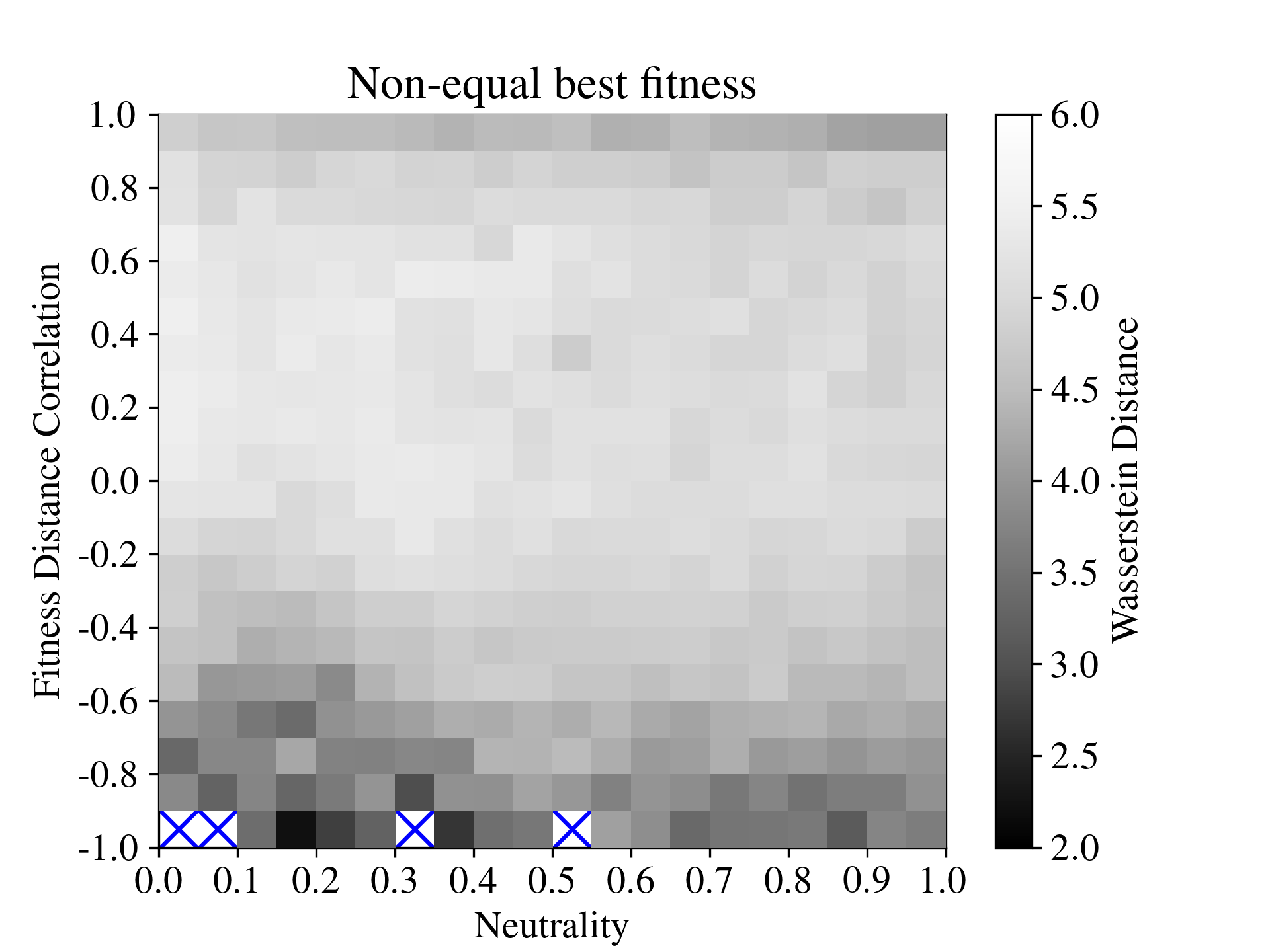}
    \includegraphics[width=.85\columnwidth]{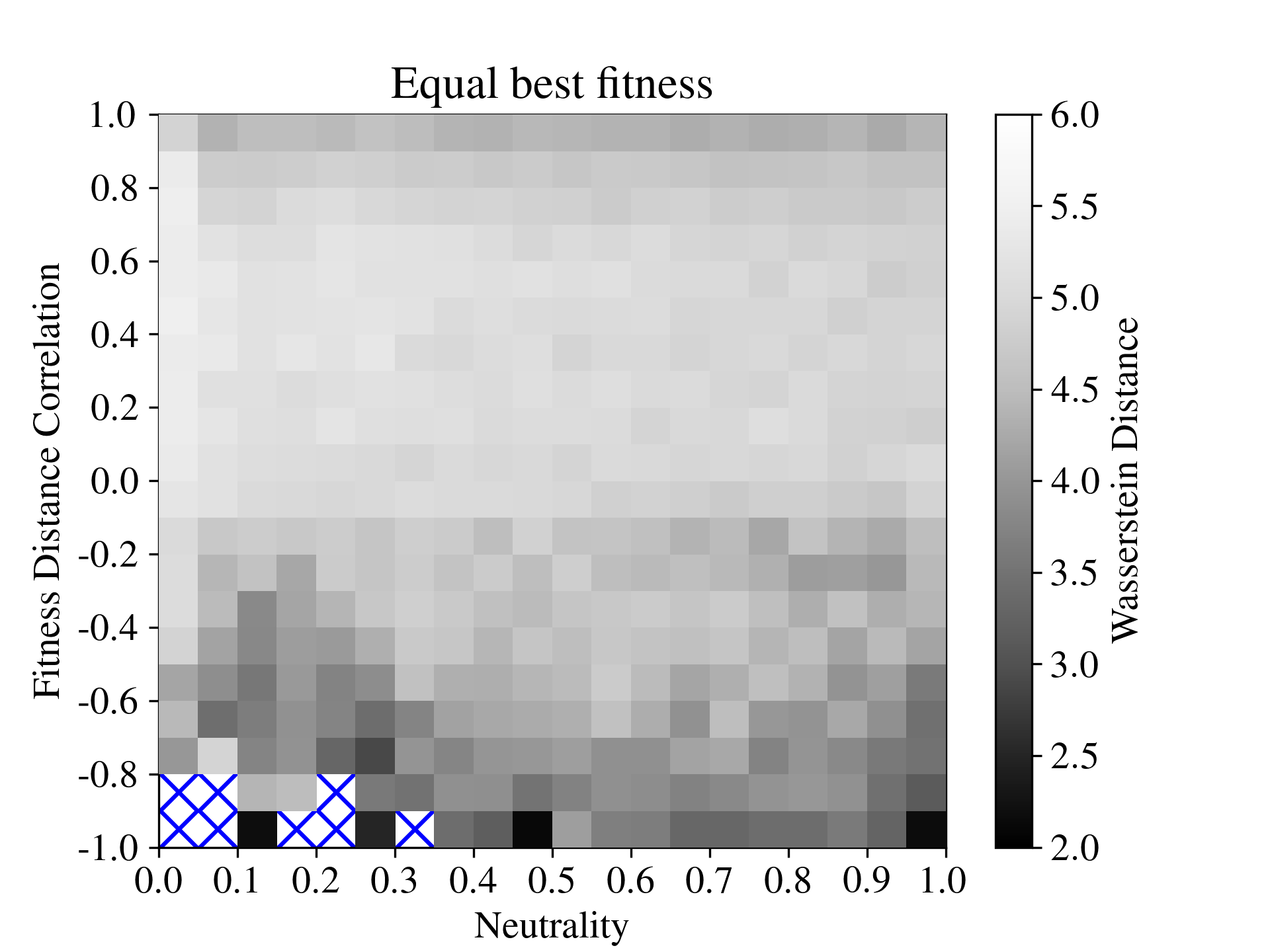}
    \caption{Average Wasserstein distance of the functions in the archive in case study 2}
    \label{fig:archive-different-algorithms}
\end{figure*}

\begin{table}[]
\caption{Train performance: $d$ between two DE configurations on our functions and CEC2005 benchmarks}
\label{tab:d-two-de}
\centering
\begin{tabular}{|c|c|c|c|}
    \hline
    \textbf{Ours} & \textbf{$d$} & \textbf{CEC} & \textbf{$d$} \\
    \hline
    $h_1$ & 5.412 & $F_9$ & 0.556 \\
    $h_2$ & 6.642 & $F_{10}$ & 0.304 \\
    $h_3$ & 6.495 & $F_{15}$ & 0.769 \\
    $h_4$ & 5.788 & $F_{16}$ & 1.494 \\
    $h_5$ & 6.289 & $F_{17}$ & 1.040 \\
    $h_6$ & 5.244 & $F_{18}$ & 0.568 \\
    $h_7$ & 5.424 & $F_{19}$ & 1.129 \\
    $h_8$ & 6.831 & $F_{20}$ & 0.894 \\
    $h_9$ & 5.422 & $F_{21}$ & 1.192 \\
    $h_{10}$ & 6.403 & $F_{22}$ & 0.914 \\
    $h_{11}$ & 5.639 & $F_{23}$ & 1.639 \\
    $h_{12}$ & 5.644 & $F_{24}$ & 0.471 \\
    $h_{13}$ & 5.714 & - & - \\
    $h_{14}$ & 6.270 & - & - \\
    $h_{15}$ & 6.262 & - & - \\
    \hline\hline
    \textbf{Ours} & \textbf{$d$} & \textbf{CEC} & \textbf{$d$} \\
    \hline
    Mean & \underline{\textbf{5.965}} & Mean & 0.914 \\
    Std. & 0.496 & Std. & 0.391 \\
    \hline
\end{tabular}
\end{table}

\begin{table}[]
\caption{Train performance: $d$ between SHADE and CMA-ES on our functions and CEC2005 benchmarks}
\label{tab:d-two-algorithms}
\centering
\begin{tabular}{|c|c|c|c|}
    \hline
    \textbf{Ours} & \textbf{$d$} & \textbf{CEC} & \textbf{$d$} \\
    \hline
    $h_{16}$ & 6.431 & $F_{9}$ & 2.306 \\
    $h_{17}$ & 6.047 & $F_{10}$ & 2.098 \\
    $h_{18}$ & 6.153 & $F_{15}$ & 3.069 \\
    $h_{19}$ & 6.024 & $F_{16}$ & 3.091 \\
    $h_{20}$ & 5.838 & $F_{17}$ & 3.181 \\
    $h_{21}$ & 6.274 & $F_{18}$ & 3.947 \\
    $h_{22}$ & 5.720 & $F_{19}$ & 3.889 \\
    $h_{23}$ & 6.391 & $F_{20}$ & 4.159 \\
    $h_{24}$ & 5.952 & $F_{21}$ & 1.815 \\
    $h_{25}$ & 5.852 & $F_{22}$ & 1.987 \\
    $h_{26}$ & 6.065 & $F_{23}$ & 2.438 \\
    $h_{27}$ & 6.143 & $F_{24}$ & 2.257 \\
    $h_{28}$ & 5.910 & - & - \\
    $h_{29}$ & 6.228 & - & - \\
    $h_{30}$ & 5.802 & - & - \\
    \hline\hline
    \textbf{Ours} & \textbf{$d$} & \textbf{CEC} & \textbf{$d$} \\
    \hline
    Mean & \underline{\textbf{6.055}} & Mean & 2.853 \\
    Std. & 0.208 & Std. & 0.786 \\
    \hline
\end{tabular}
\end{table}

In this first case study, we evolve benchmark functions that maximize the difference between two parametrization of Differential Evolution (DE)~\cite{storn1997differential}. 
%
The differential mutation in DE is as in (\ref{eq:differential-mutation}). After the differential mutation, DE performs a binary crossover between the child and the parent with probability $\mathrm{CR}$.
\begin{gather}
\label{eq:differential-mutation}
    \mathbf{y} = \mathbf{x}_1 + F \cdot (\mathbf{x}_2-\mathbf{x}_3)
\end{gather}

DE is well known for being very sensitive to the values of its parameters $F$ and $\mathrm{CR}$, so we use it as a proof of concept of our proposed method, with the two sets of values below. For other parameters, we set the population size of DE as 20 and the max generation as 25.

\begin{itemize}
    \item $F=0.5$, $\mathrm{CR}=0.9$
    \item $F=0.3$, $\mathrm{CR}=0.9$
\end{itemize}

\subsubsection{Training performance}

The heatmap in Fig.~\ref{fig:archive-de-parameterization} illustrates the average Wasserstein distance of the archive in 15 runs. The blue cross indicates that no solutions are found in that cell. The left subplot shows the functions where two DE hold different best fitness values, while the right subplot shows the functions where the best fitness values of the two DE are equal. Functions with FDC and neutrality near 0 have higher Wasserstein distance values, i.e., larger difference between two DE parameter settings. This observation may indicate that these two configurations of DE are more different when FDC and neutrality are close to 0.

Table~\ref{tab:d-two-de} shows the largest Wasserstein distance (with different best fitness values) found between two configuration in 15 runs and the same distance measure on the two-dimensional version of 12 functions from 25 in CEC2005 benchmarks~\cite{suganthan2005problem} with boundary constraint of ${[-5,5]}$ (since our functions have the same search spaces). The benchmark functions generated by our method differentiate the two settings of DE better than the CEC2005 benchmark functions in the training phase (with statistical significance of $\alpha=0.05$ in a Wilcoxon test).

\subsubsection{Test performance}

To validate the ability of the benchmark set generated by the proposed GP to differentiate algorithms in a more general context, we further compare the differences of the algorithms using the same functions that were generated in the last experiment, but with higher dimensions and longer running times. We select functions generated by GP with the largest Wasserstein distances and the non-equal best fitness for two DE configurations for each run. We create 10-dimensional functions $h^{10}$ based on these functions $h$ using the following approach:
\begin{equation}
\label{eq:10-dimensional-function}
    h^{10}(x_1,x_2,\dots,x_{10}) =
    \frac{1}{9} \Sigma_{i=1}^9 h(x_i,x_{i+1})
\end{equation}

We compare our functions with the 12 CEC2005 benchmarks \cite{suganthan2005problem} and check how they differentiate the two configuration of DE algorithms in terms of best decision variables and best fitness values. This time, we run DE algorithms with 21 repetitions for each function. Different from the previous experiment, every single run costs 100000 evaluations. We use the following two metrics in (\ref{eq:delta-x}) and (\ref{eq:delta-f}). $\Delta_x$ is the average distance between two sets of the best solutions in 21 repetitions. $\mathrm{UB}$ and $\mathrm{LB}$ are the boundaries of the search space. $D$ is the dimension of the problem. $\Delta_f$ is the average scaled difference in fitness values between two best solution sets, where $f^{\max}$ and $f^{\min}$ are the maximum and minimum fitness values of the solutions in $\mathcal{A}\cup\mathcal{B}$, respectively. For both metrics, the larger value indicates a better performance of the function in differentiating two algorithms/configurations.
\begin{equation}
\label{eq:delta-x}
    \Delta_x = \frac{1}{|\mathcal{A}| \cdot |\mathcal{B}|} \frac{\Sigma_{\mathbf{a}\in\mathcal{A}} \Sigma_{\mathbf{b}\in\mathcal{B}} ||\mathbf{a} - \mathbf{b}||}{\sqrt{\mathrm{D}}(\mathrm{UB}-\mathrm{LB})}
\end{equation}

\begin{equation}
\label{eq:delta-f}
    \Delta_f = \frac{1}{|\mathcal{A}| \cdot |\mathcal{B}|} \frac{\Sigma_{\mathbf{a}\in\mathcal{A}} \Sigma_{\mathbf{b}\in\mathcal{B}} |f(\mathbf{a}) - f(\mathbf{b})|}{f^{\max}-f^{\min}}
\end{equation}

As shown in Table~\ref{tab:metric-two-de}, functions generated by GP hold a larger mean value of $\Delta_x$, and thus differentiate the two DE algorithms better in terms of best decision variables  with a statistical significance (i.e., the underlined mean value shows the p-value of the Wilcoxon ranked sum test is less than 0.05). On the other hand, functions in CEC2005 benchmark differentiate the two settings better in the fitness value (larger $\Delta_f$), however, without statistical significance.

\subsection{Case Study on Different Algorithms}
\label{sec:case-study-on-different-algorithms}

In our second case study, we focus on finding a function to differentiate two powerful evolutionary optimizers, namely Success-History based Adaptive Differential Evolution (SHADE)~\cite{tanabe2013success} and Covariance Matrix Adaption Evolutionary Strategy (CMA-ES)~\cite{hansen2001completely}. 

SHADE~\cite{tanabe2013success} generates parameters of differential mutation for every individual based on Cauchy distribution and Gaussian distribution. The parameters that lead to successful mutations in one generation are used to estimate the expectation values for parameter distributions. SHADE memorizes the estimated expectation values in the last $H$ generations. CMA-ES~\cite{hansen2001completely} samples individuals based on a multivariate normal distribution and updates the expectation vector and covariance matrix based on elite solutions.


We first run our GP algorithm to generate a set of functions using the same approach as in Section~\ref{sec:case-study-on-parameterization-of-differential-evolution}. In particular, we compare the following two settings of SHADE and CMA-ES. For both algorithm, we use a maximum number of evaluations of 500. The population size of SHADE and CMA-ES is set to 20. Some important parameters are as follows. The reader can refer to the original works of the algorithms~\cite{hansen2001completely,tanabe2013success} for the details of the parameters.

\begin{itemize}
    \item SHADE: $H=20$, $p_{\max}=0.2$
    \item CMA-ES: $\sigma=6$
\end{itemize}

\subsubsection{Training performance}

In Fig.~\ref{fig:archive-different-algorithms}, functions with positive FDC hold large Wasserstein distance, and thus show the difference between the two algorithms well. Moreover, it is easier to find a function to differentiate SHADE and CMA-ES, compared to differentiating the two DE parameter configuration (i.e., FDC larger than -0.2). This observation is natural, since SHADE and CMA-ES work in quite different manners (though both of them are strong optimizers) as described in the beginning of this section. We compare the 15 functions generated in all repetitions with the functions in CEC2005 benchmark. As shown in Table~\ref{tab:d-two-algorithms}, our functions show significantly larger difference between SHADE and CMA-ES, compared with CEC2005 benchmark functions.

\begin{figure*}
\centering
\includegraphics[width=.187\textwidth]{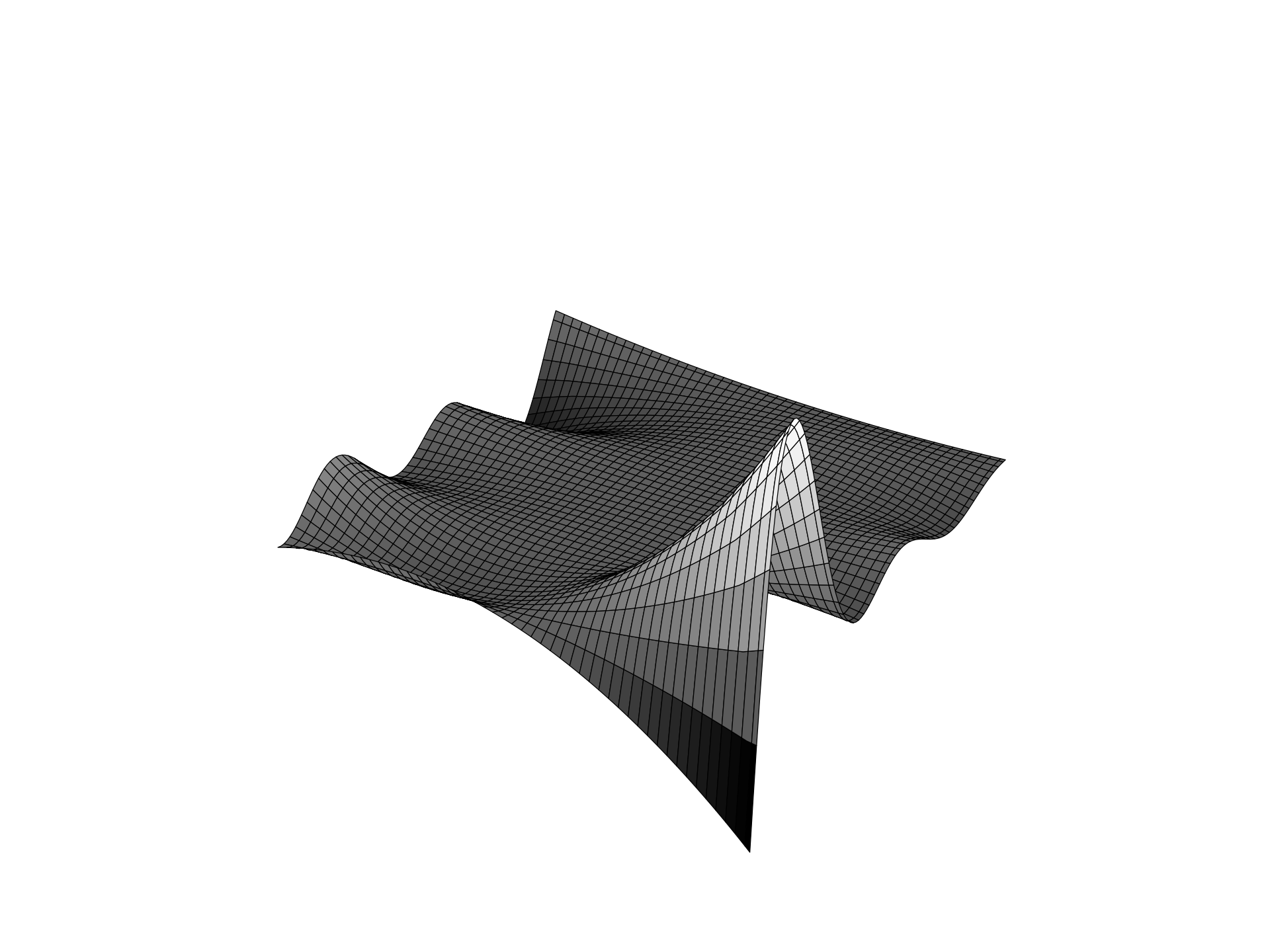}
\includegraphics[width=.187\textwidth]{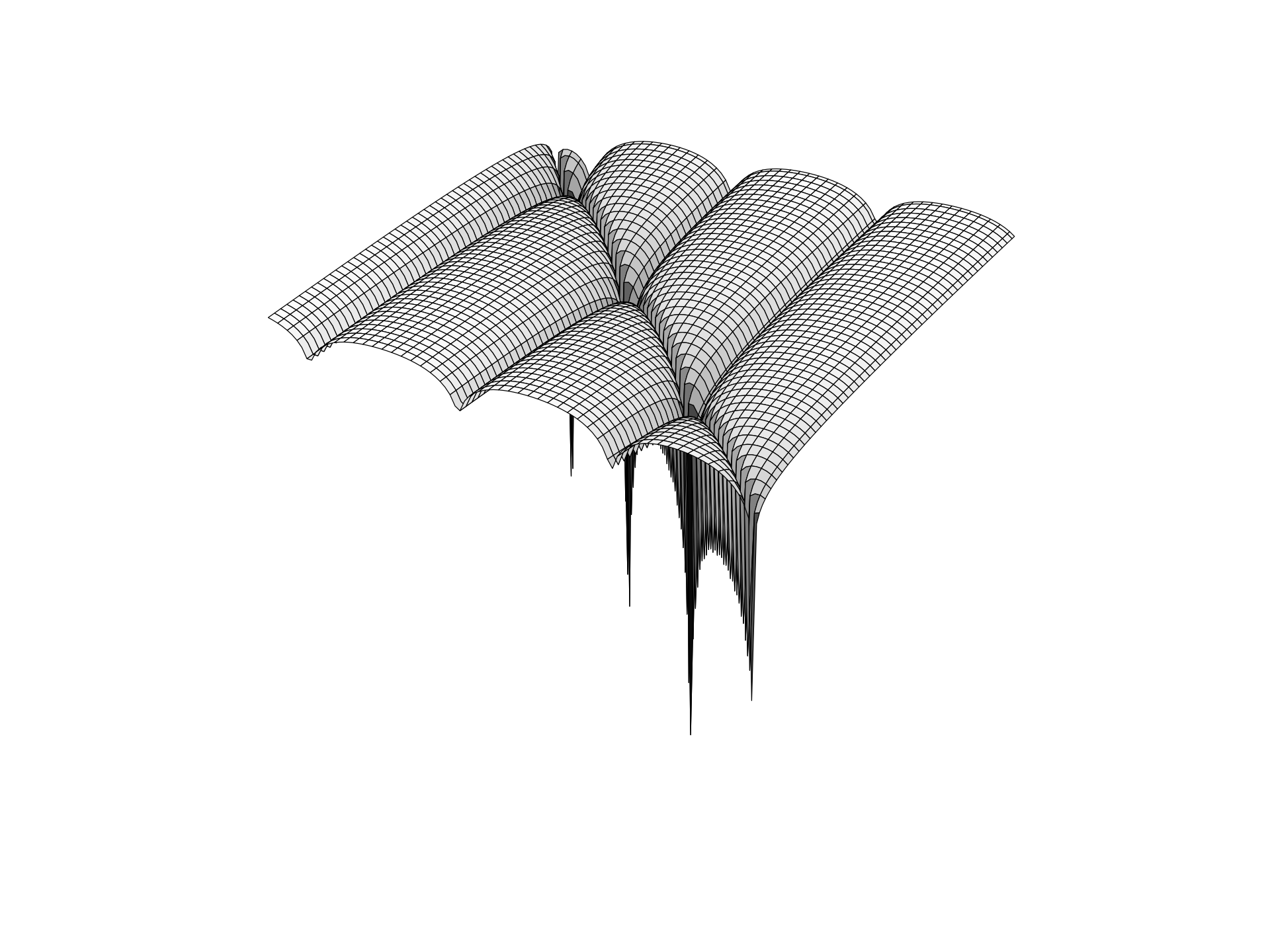}
\includegraphics[width=.187\textwidth]{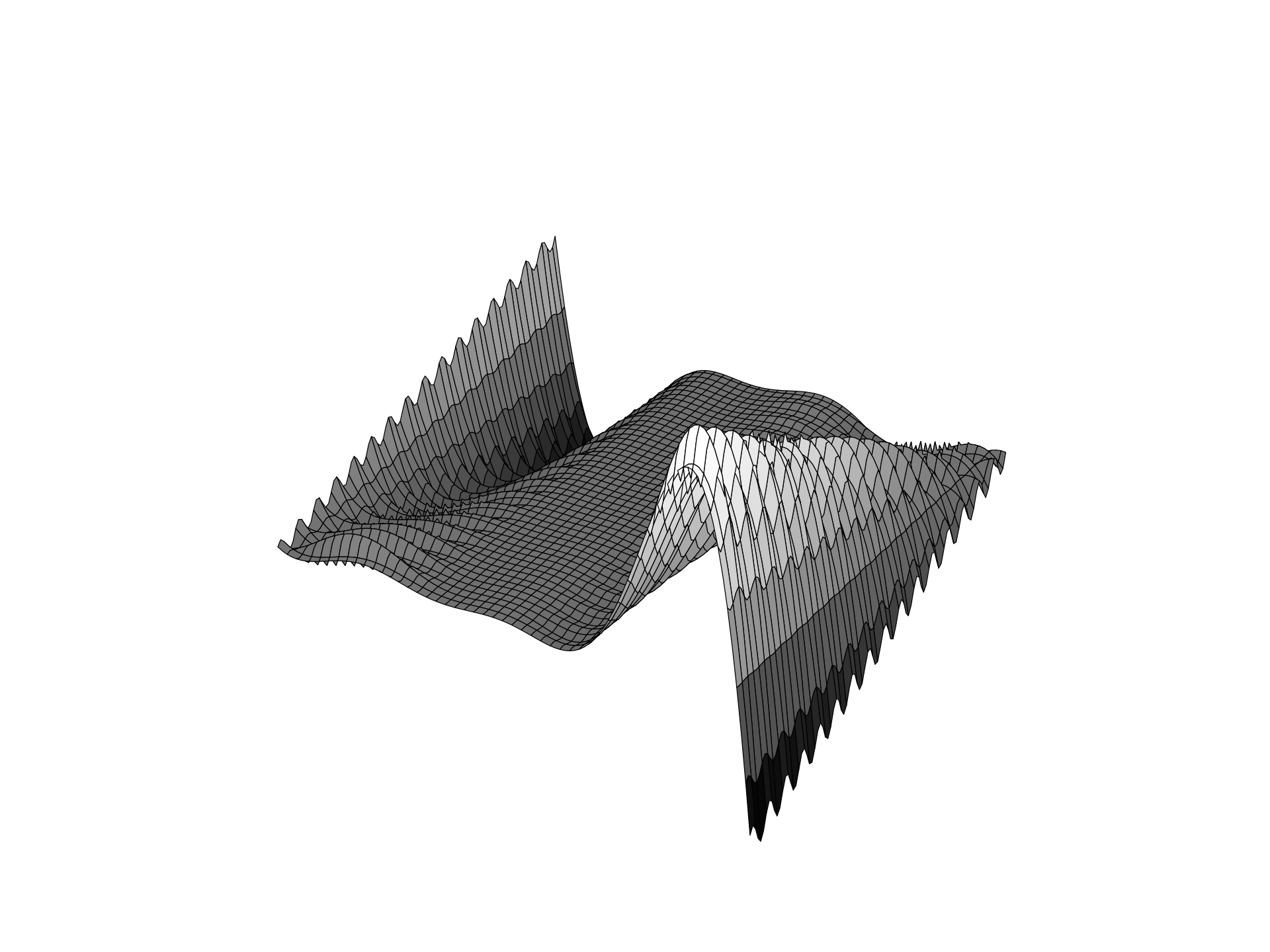}
\includegraphics[width=.187\textwidth]{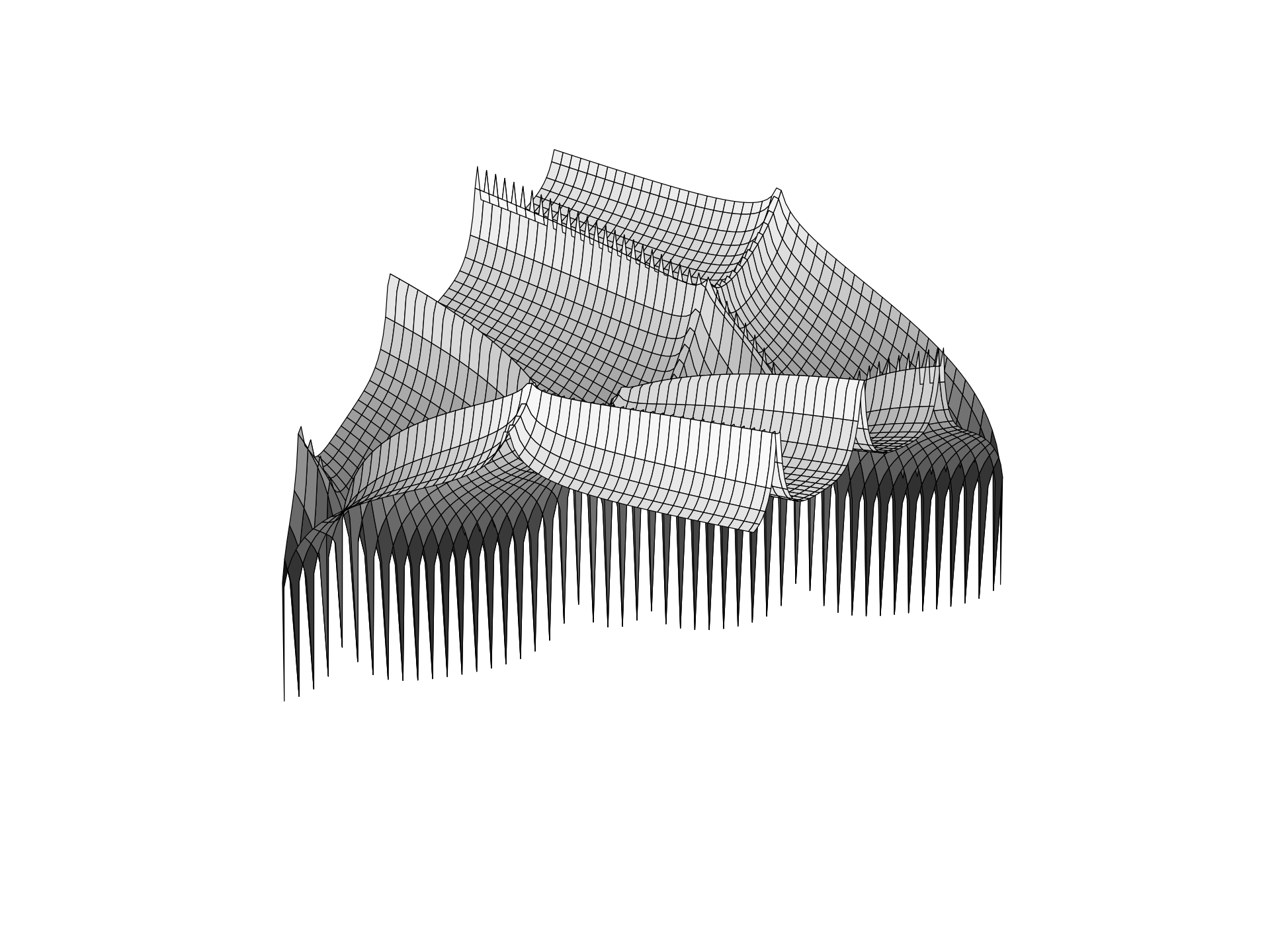}
\includegraphics[width=.187\textwidth]{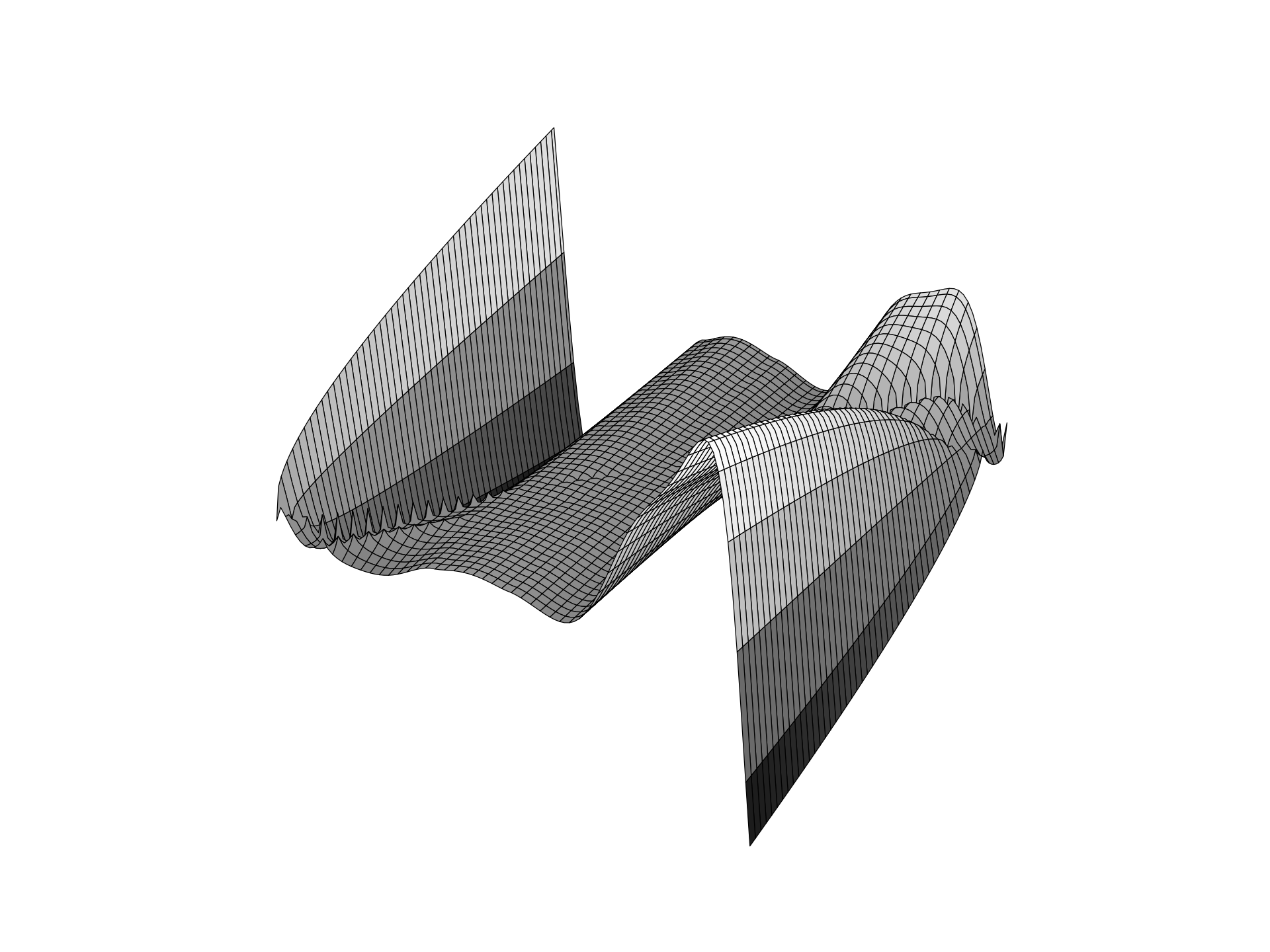}\\
\includegraphics[width=.187\textwidth]{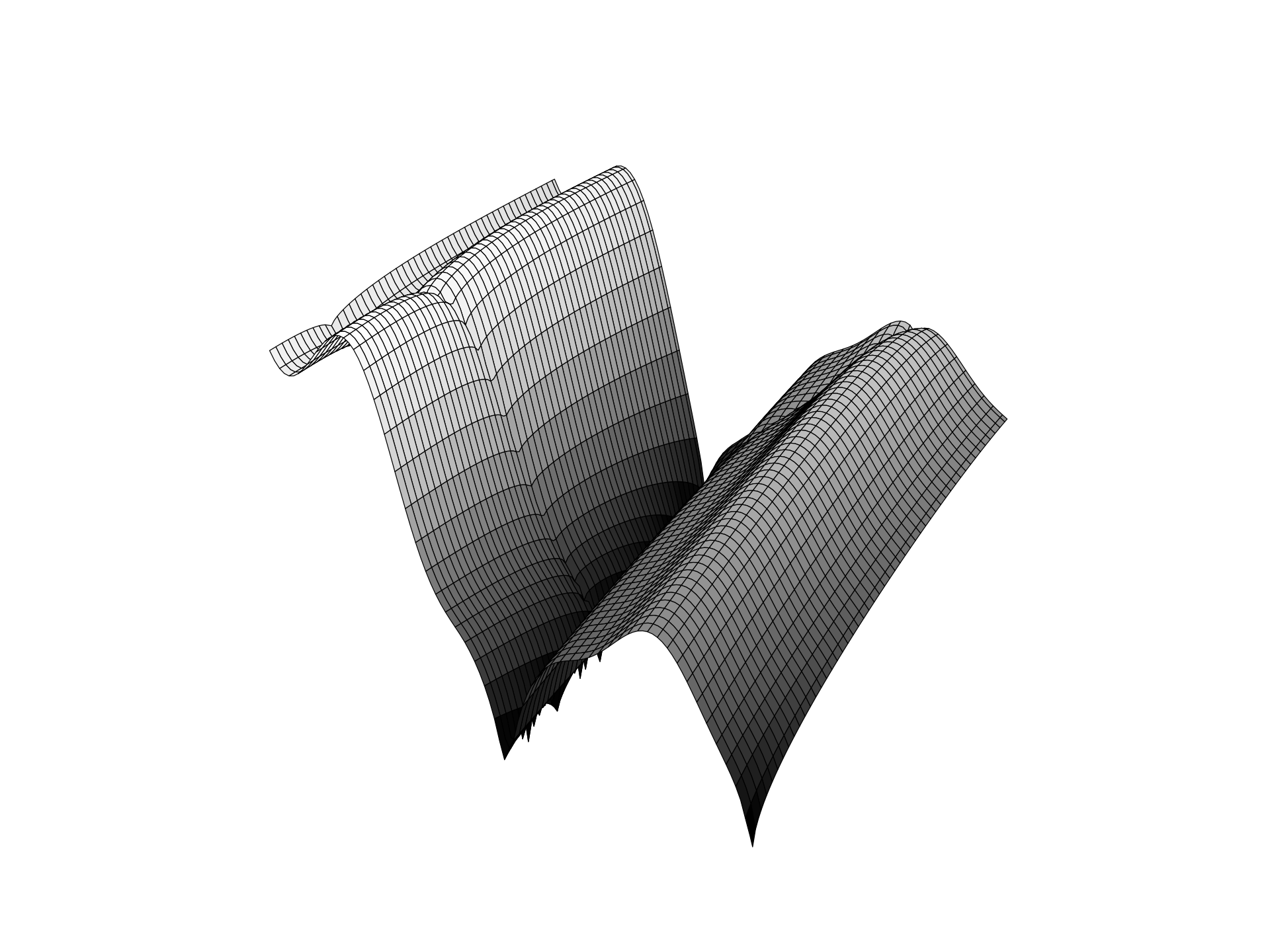}
\includegraphics[width=.187\textwidth]{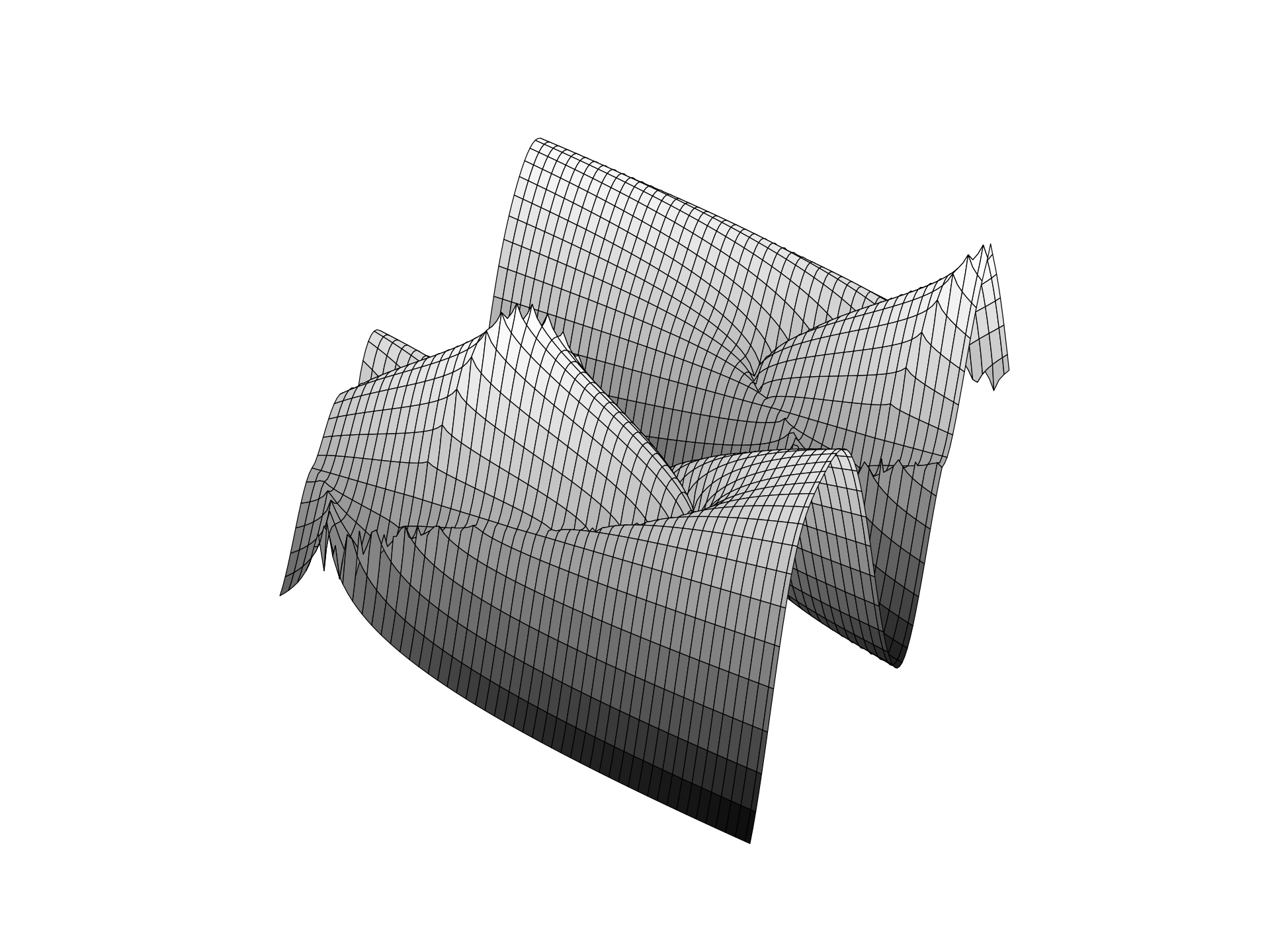}
\includegraphics[width=.187\textwidth]{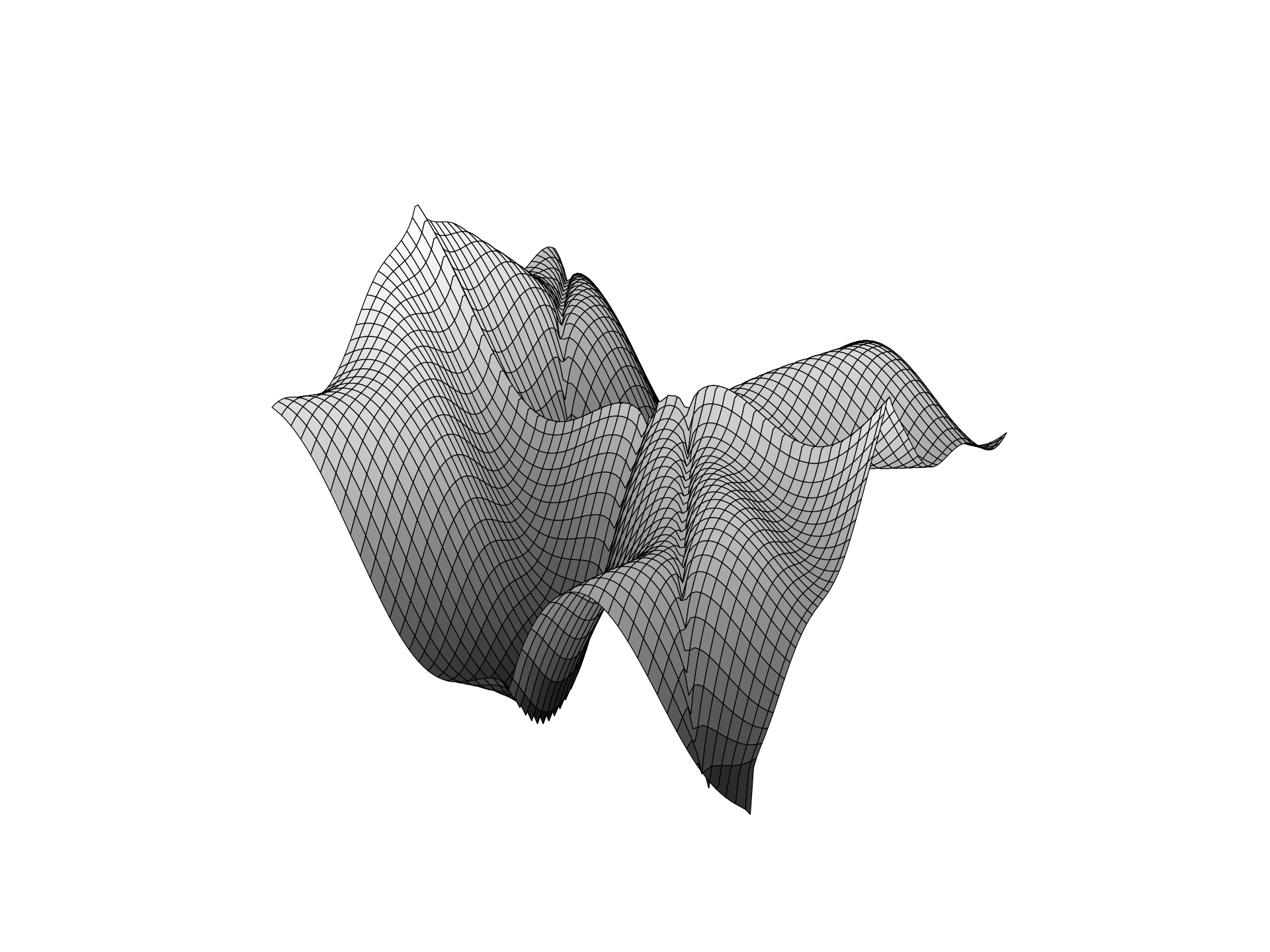}
\includegraphics[width=.187\textwidth]{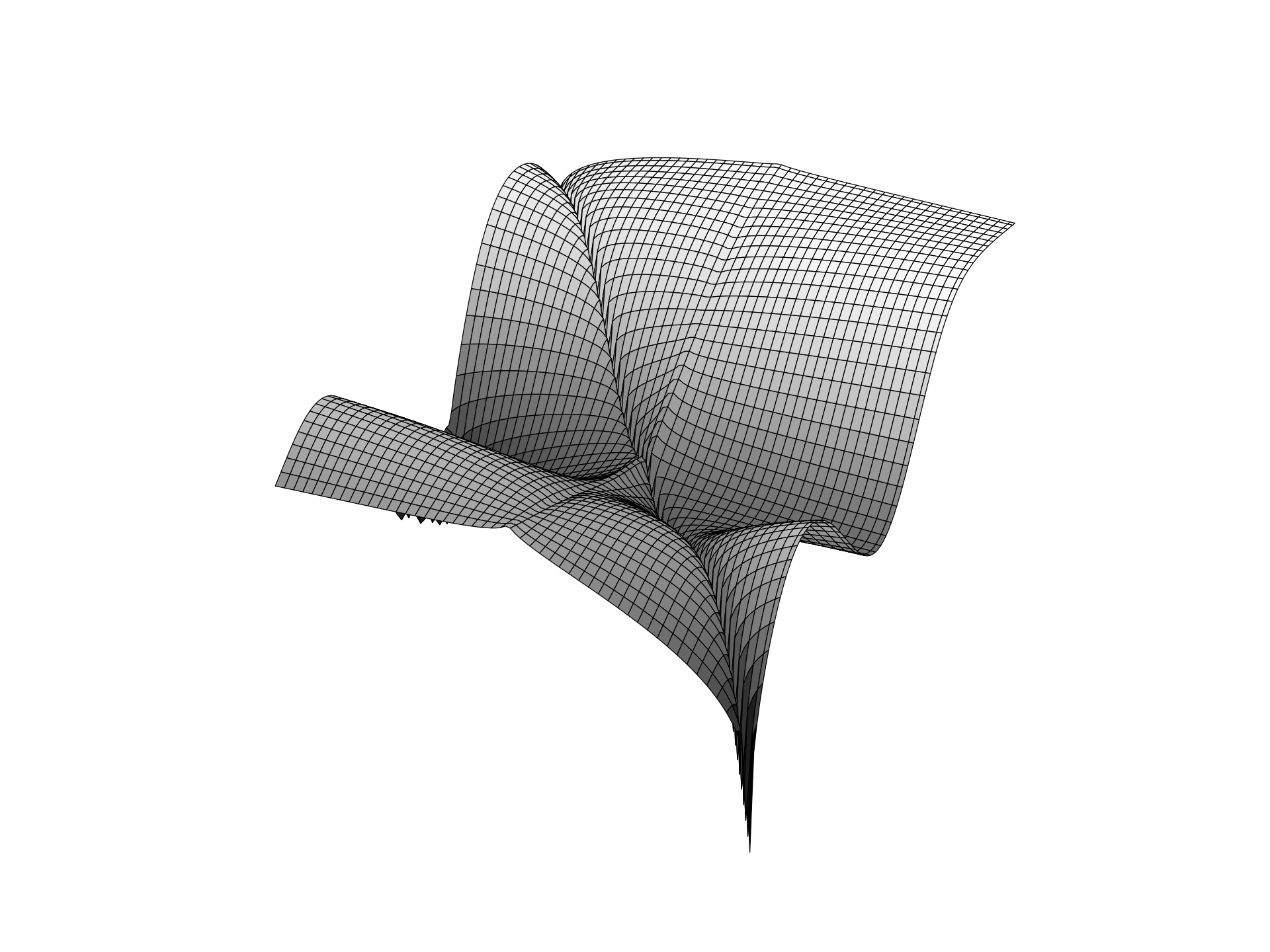}
\includegraphics[width=.187\textwidth]{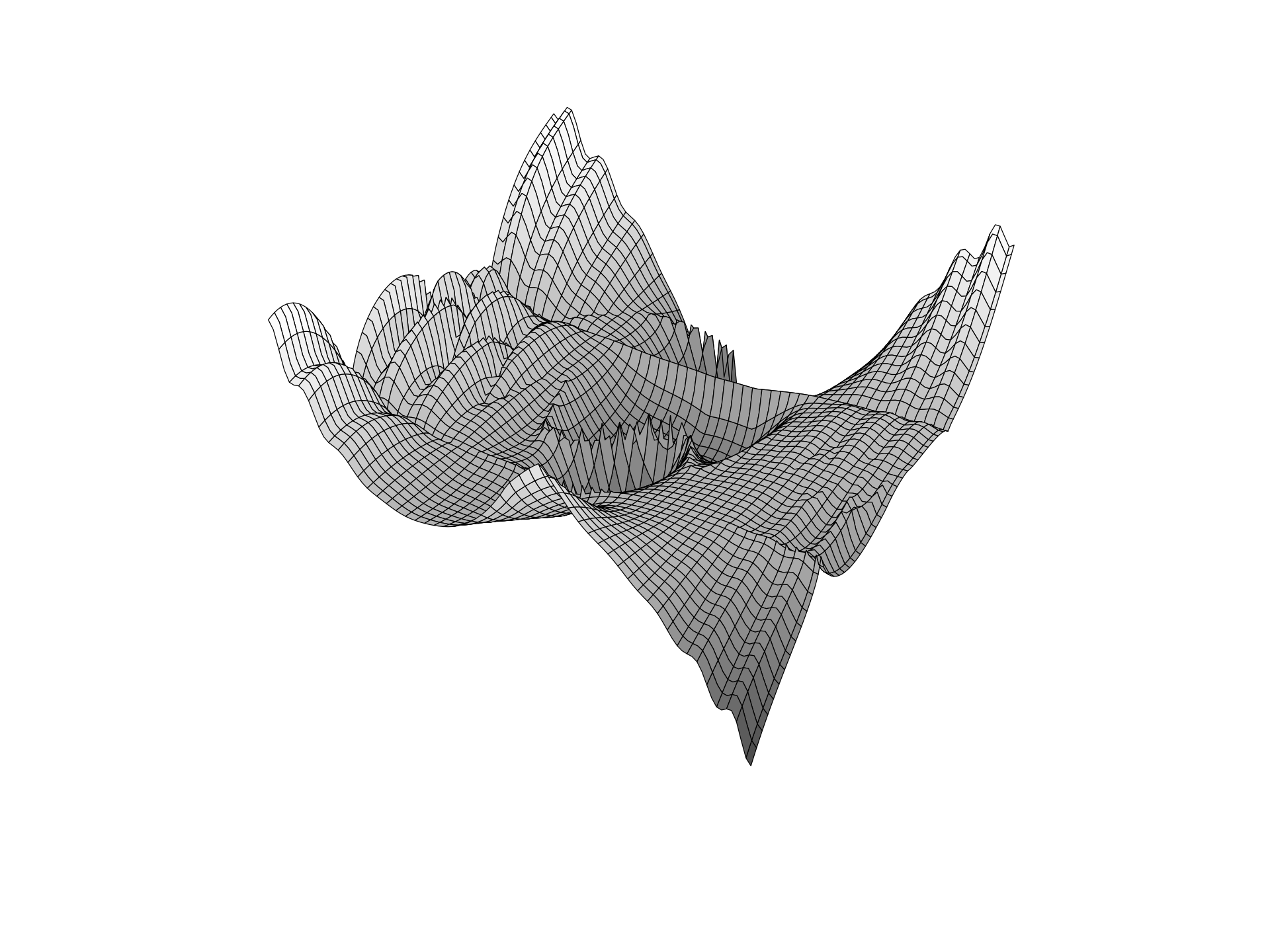}
\caption{Functions with Top-5 Wasserstein distance in the two case studies (first line: case study 1, second line: case study 2)}
\label{fig:function-set}
\end{figure*}

\subsubsection{Test performance}

\begin{table}[]
\caption{Test performance: $\Delta_x$ and $\Delta_f$ of our functions and CEC2005 benchmarks on differentiating two DE configurations}
\label{tab:metric-two-de}
\centering
\begin{tabular}{|c|c|c|c|c|c|}
    \hline
    \textbf{Ours} & \textbf{$\Delta_x$} & \textbf{$\Delta_f$} & \textbf{CEC} & \textbf{$\Delta_x$} & \textbf{$\Delta_f$} \\
    \hline
    $h_1$    & 0.607 & 0.336  & $F_9$    & 0.092 & 0.241 \\
    $h_2$    & 0.638 & 0.278  & $F_{10}$ & 0.095 & 0.294 \\
    $h_3$    & 0.589 & 0.288  & $F_{15}$ & 0.217 & 0.222 \\
    $h_4$    & 0.696 & 0.213  & $F_{16}$ & 0.189 & 0.264 \\
    $h_5$    & 0.587 & 0.338  & $F_{17}$ & 0.193 & 0.291 \\
    $h_6$    & 0.212 & 0.486  & $F_{18}$ & 0.332 & 0.342 \\
    $h_7$    & 0.561 & 0.299  & $F_{19}$ & 0.335 & 0.325 \\
    $h_8$    & 0.586 & 0.327  & $F_{20}$ & 0.322 & 0.309 \\
    $h_9$    & 0.641 & 0.178  & $F_{21}$ & 0.208 & 0.361 \\
    $h_{10}$ & 0.633 & 0.330  & $F_{22}$ & 0.253 & 0.320 \\
    $h_{11}$ & 0.658 & 0.291  & $F_{23}$ & 0.172 & 0.466 \\
    $h_{12}$ & 0.612 & 0.246  & $F_{24}$ & 0.190 & 0.423 \\
    $h_{13}$ & 0.651 & 0.376  & - & - & -\\
    $h_{14}$ & 0.595 & 0.284  & - & - & -\\
    $h_{15}$ & 0.607 & 0.315  & - & - & -\\
    \hline\hline
    \textbf{Ours} & \textbf{$\Delta_x$} & \textbf{$\Delta_f$} & \textbf{CEC} & \textbf{$\Delta_x$} & \textbf{$\Delta_f$} \\
    \hline
    Mean & \underline{\textbf{0.592}} & 0.306 & Mean & 0.217 & \textbf{0.322} \\
    Std. & 0.107 & 0.069 & Std. & 0.079 & 0.068 \\
    \hline
\end{tabular}
\end{table}

\begin{table}[]
\caption{Test performance: $\Delta_x$ and $\Delta_f$ of our functions and CEC2005 benchmarks on differentiating SHADE and CMA-ES}
\label{tab:metric-two-algorithms}
\centering
\begin{tabular}{|c|c|c|c|c|c|}
    \hline
    \textbf{Ours} & \textbf{$\Delta_x$} & \textbf{$\Delta_f$} & \textbf{CEC} & \textbf{$\Delta_x$} & \textbf{$\Delta_f$} \\
    \hline
    $h_{16}$ & 0.085 & 0.616 & $F_9$    & 0.029 & 0.238 \\
    $h_{17}$ & 0.541 & 0.886 & $F_{10}$ & 0.088 & 0.631 \\
    $h_{18}$ & 0.372 & 0.319 & $F_{15}$ & 0.273 & 0.682 \\
    $h_{19}$ & 0.002 & 0.769 & $F_{16}$ & 0.053 & 0.354 \\
    $h_{20}$ & 0.202 & 0.342 & $F_{17}$ & 0.058 & 0.590 \\
    $h_{21}$ & 0.374 & 0.353 & $F_{18}$ & 0.228 & 0.501 \\
    $h_{22}$ & 0.499 & 0.254 & $F_{19}$ & 0.341 & 0.379 \\
    $h_{23}$ & 0.455 & 0.260 & $F_{20}$ & 0.276 & 0.437 \\
    $h_{24}$ & 0.229 & 0.118 & $F_{21}$ & 0.201 & 0.130 \\
    $h_{25}$ & 0.496 & 0.216 & $F_{22}$ & 0.141 & 0.303 \\
    $h_{26}$ & 0.045 & 0.147 & $F_{23}$ & 0.281 & 0.117 \\
    $h_{27}$ & 0.463 & 0.701 & $F_{24}$ & 0.007 & 0.245 \\
    $h_{28}$ & 0.394 & 0.395 & - & - & -\\
    $h_{29}$ & 0.494 & 0.397 & - & - & -\\
    $h_{30}$ & 0.334 & 0.481 & - & - & -\\
    \hline\hline
    \textbf{Ours} & \textbf{$\Delta_x$} & \textbf{$\Delta_f$} & \textbf{CEC} & \textbf{$\Delta_x$} & \textbf{$\Delta_f$} \\
    \hline
    Mean & \underline{\textbf{0.332}} & \textbf{0.417} & Mean & 0.165 & 0.384 \\
    Std. & 0.172 & 0.222 & Std. & 0.111 & 0.181 \\
    \hline
\end{tabular}
\end{table}

The test performance of this case study is shown in Table~\ref{tab:metric-two-algorithms}. To calculate the test performance, we run algorithms with 100000 evaluations. The population size and $H$ parameter of SHADE are set to 100, while the population size of CMA-ES is 200. Our functions produce larger average difference in both decision space and fitness value measure. Especially in the decision space, our functions hold better $\Delta_x$ with statistical significance. These data indicates that our functions are better to highlight the difference between SHADE and CMA-ES.

\section{Discussion on the Generated Benchmarks}
\label{sec:discussion-on-the-generated-benchmarks}

We further show the details of the generated benchmark functions in this section. Fig.~\ref{fig:function-set} shows the functions with Top-5 Wasserstein distances in both studies. The 3D surface plots and the mathematical expression of all functions ($h_1$--$h_{30}$) can be found at \url{https://github.com/Y1fanHE/cec2024}.

\begin{figure*}
\centering
\subfigure[$h_8$]{
    \includegraphics[width=.85\columnwidth]{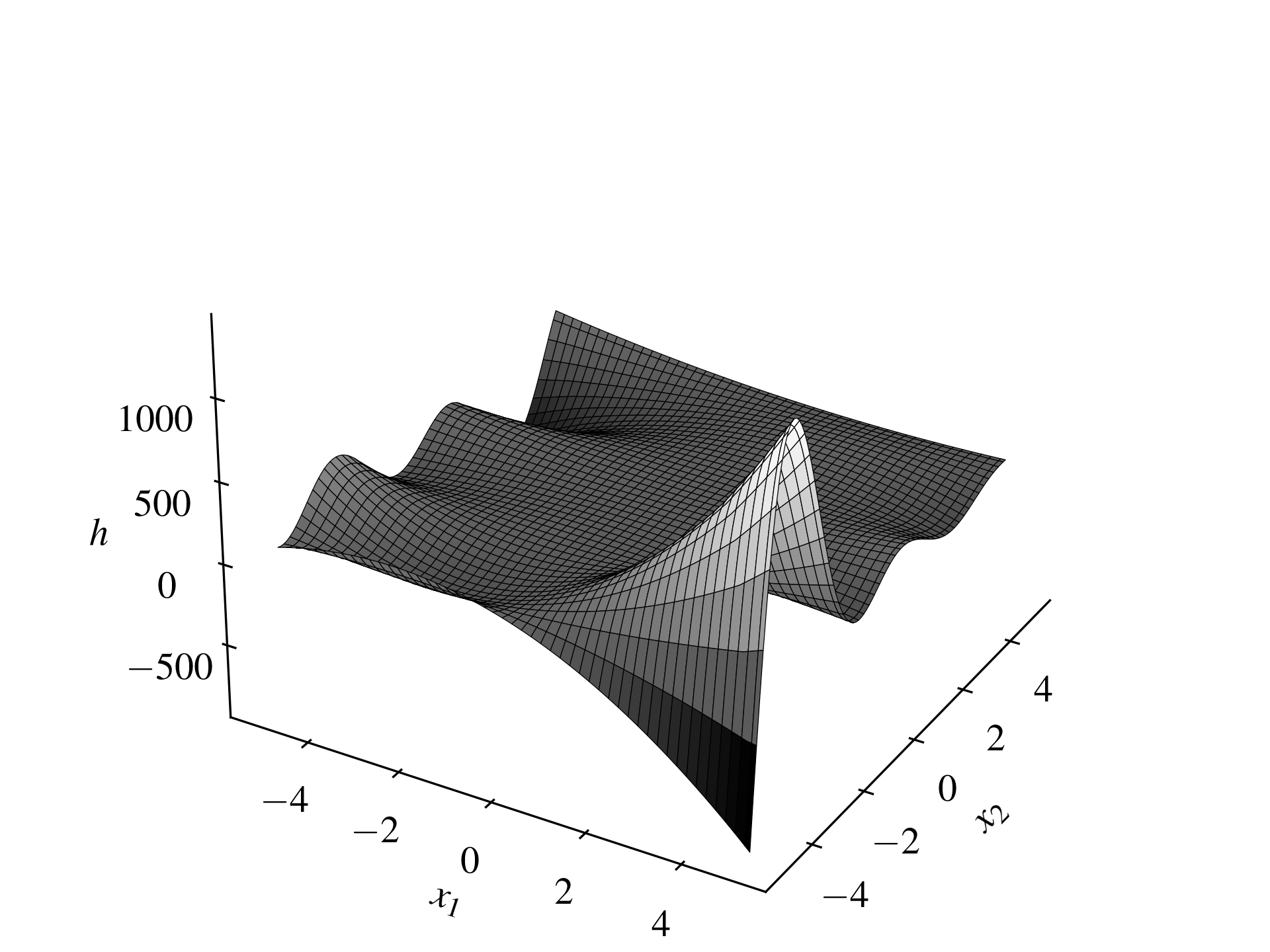}}
\subfigure[$h_{16}$]{
    \includegraphics[width=.85\columnwidth]{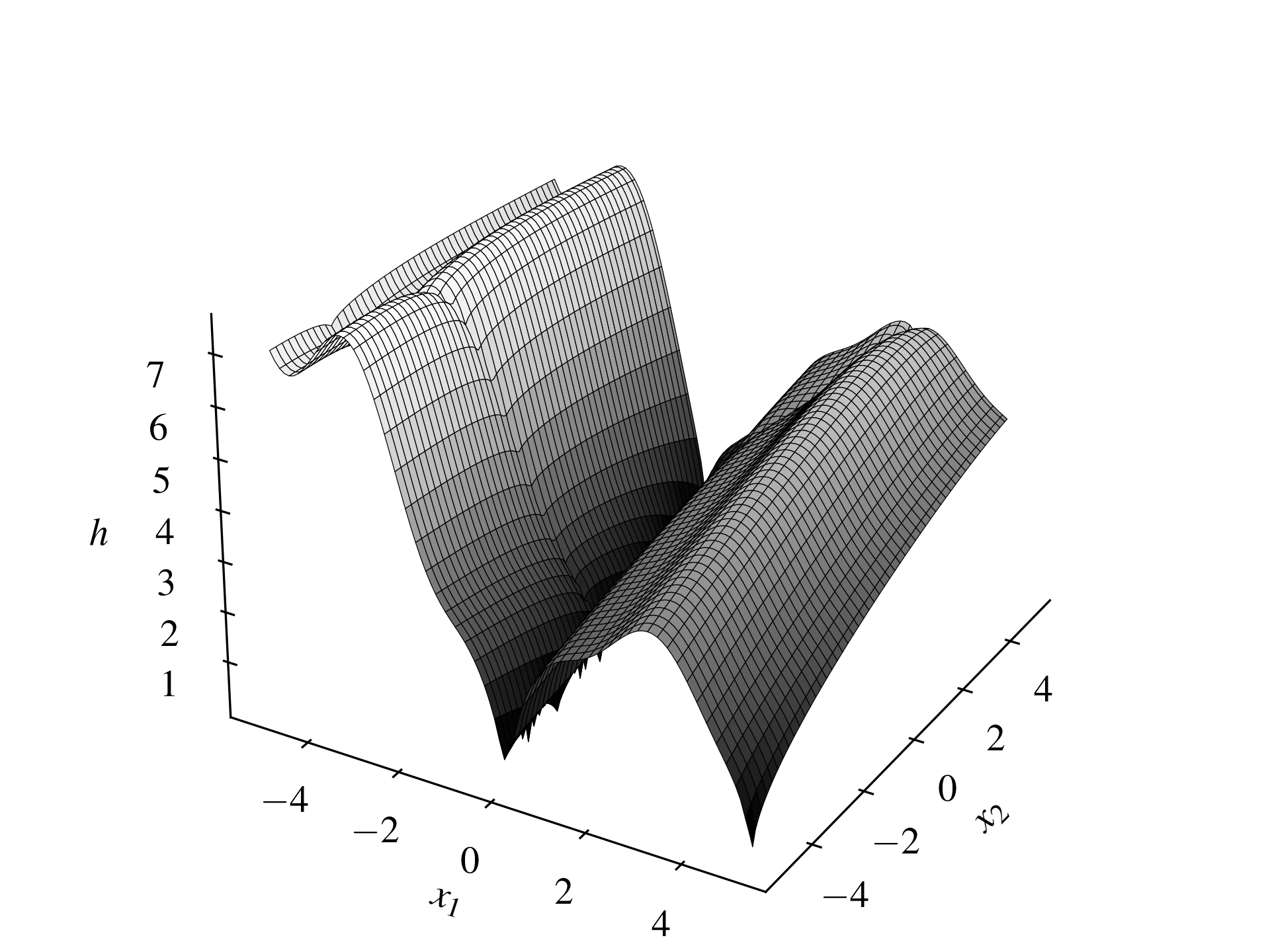}}
\caption{3D surface plots of the best functions generated by GP in the case studies}
\label{fig:3d-surface}
\end{figure*}

\begin{figure*}[!h]
\centering
\subfigure[$h_8$]{
    \includegraphics[width=.85\columnwidth]{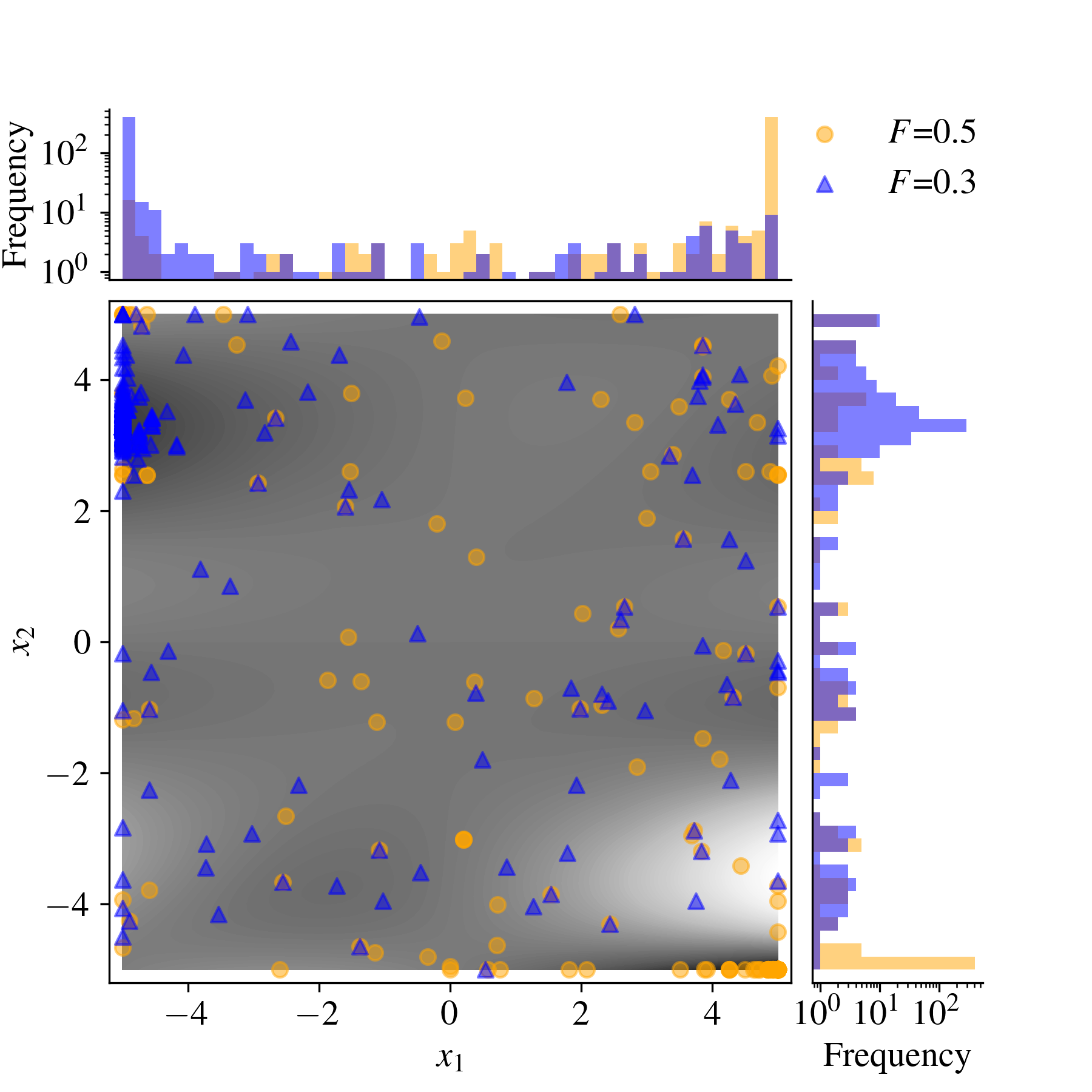}}
\subfigure[$h_{16}$]{
    \includegraphics[width=.85\columnwidth]{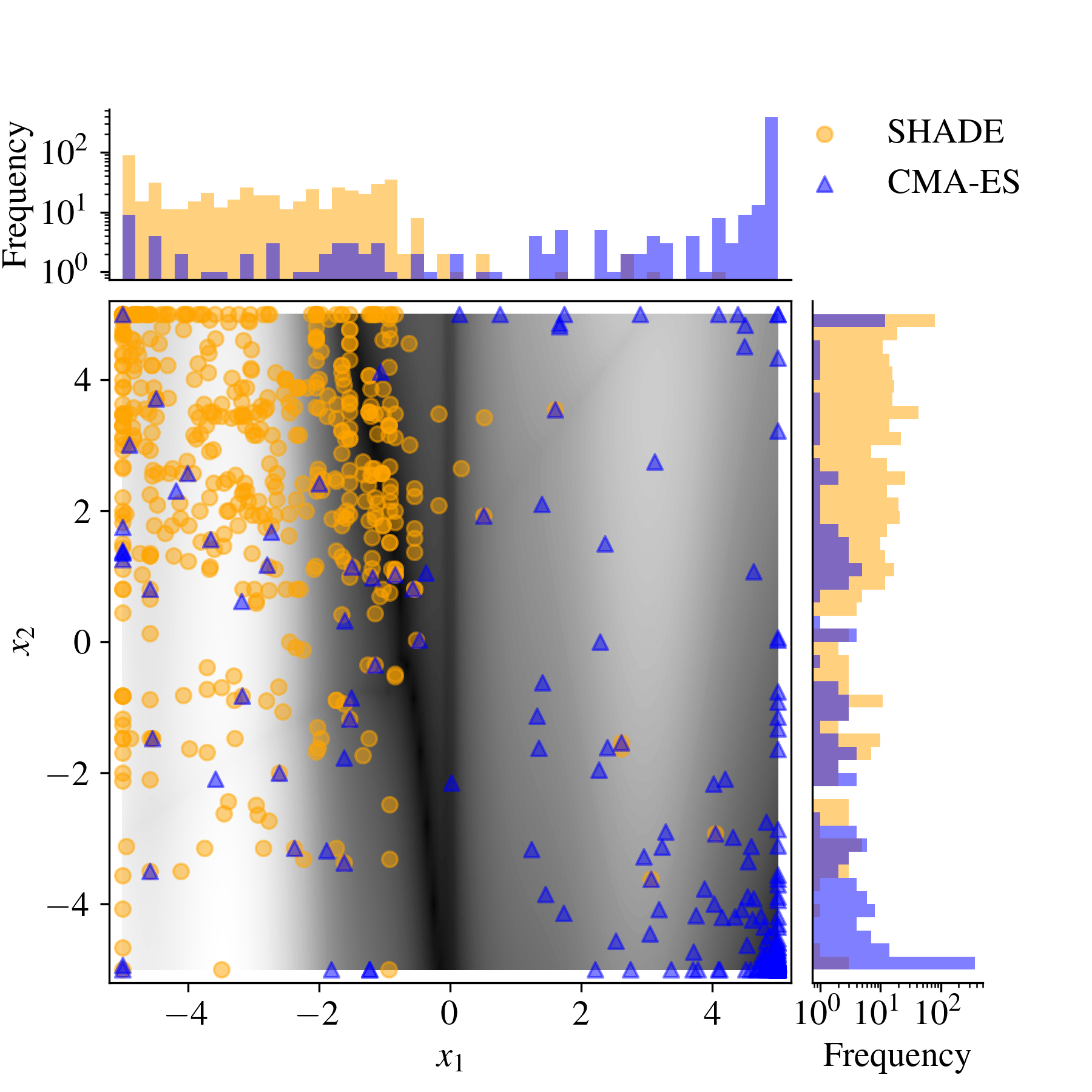}}
\caption{Scatter plots and histogram plots (frequency in log scale) of the solutions sampled in the training phase}
\label{fig:scatter-plot}
\end{figure*}

For both case studies, we select the function that hold largest Wasserstein distance and different best fitness values between two algorithms in the training phase. They are $h_8$ for the first case study and $h_{16}$ for the second case study. The 3D surface plots of the two functions are in Fig.~\ref{fig:3d-surface}. The solutions generated in 500 evaluations (same parameters as in Section~\ref{sec:experiments}) on the two functions are shown in Fig.~\ref{fig:scatter-plot}. In the same figures, we additionally plot the parameter distributions of the solution sets with the frequency in log scale.

$h_8$ has multiple local optimums and a large area of plateaus. Some local optimums are located at the edges of the search space. When using $F=0.5$, the DE algorithm has a larger mutation range, and thus the optimizer has more chance to explore and reach the global optimum at $(5,-5)$ compared with $F=0.3$. $h_{16}$ differentiates between SHADE and CMA-ES. It contains multiple local optimums as well. SHADE returns a solution set that is widely spread on the search space, while the solutions produced by CMA-ES are more concentrated near $(5,-5)$. Moreover, it is illustrated in the histogram that SHADE failed to explore the area near $x_1=5$.

\section{Conclusions}
\label{sec:conclusions}

In this study, we investigate how to automatically compose benchmark functions to differentiate between a pair of evolutionary optimizers. Our proposed method is to use Genetic Programming (GP) as an initial approach. We use the Wasserstein distance of the parameter distribution of the solution sets of both optimizers as the evaluation metric for evolving the benchmark function. Additionally, we introduced MAP-Elites using landscape metrics to enhance the performance of GP, and illustrate how the algorithms are different in varying landscape characteristics.

We compared the functions generated by our GP with the functions from CEC2005 benchmark. 
The comparison included an analysis of the functions generated directly by the GP (train performance) as well as an extended validation of the functions using an increased amount of evaluations and dimensions than those used in the GP (test performance).

Our functions could differentiate better in the decision space between a pair of DE configurations ($F=0.3$ and $F=0.5$) as well as a pair of powerful optimizers (SHADE and CMA-ES). We illustrated how the algorithms are different under various fitness distance correlation (FDC) and neutrality of the functions. Our method is promising to compose new optimization functions to comparing evolutionary algorithms.

In the future, we plan to introduce distance measure in the fitness space into the current evaluation metric to make more difference between optimizers in objective values. Moreover. our initial design of the MAP-Elites used FDC and neutrality as phenotypic descriptor. One of our future work is to explore other possible indicators for the phenotypical diversity in MAP-Elites, that might produce more interesting functions. We expect that our GP-based benchmark generator will be a useful component of hyper-heuristic systems.



\bibliographystyle{ieeetr}
\bibliography{ref}

\end{document}